\documentclass[conference]{IEEEtran}
\IEEEoverridecommandlockouts
\usepackage{amsmath,amssymb,amsfonts}
\usepackage{graphicx}
\usepackage{fancyhdr}
\usepackage{amsfonts}
\usepackage{textcomp}
\usepackage{xcolor}
\usepackage{pifont}
\usepackage{float}
\usepackage{subfig}
\usepackage{bm}
\usepackage{makecell}
\usepackage[noend]{algorithmic}
\usepackage{algorithm}
\usepackage[normalem]{ulem}
\usepackage[hyphens]{url}
\usepackage{cite}

\newcommand{\pluseq}{\mathrel{{+}{=}}}
\newcommand{\minuseq}{\mathrel{{-}{=}}}

\AtBeginDocument{%
  }

\fancypagestyle{firstpage}{
  \fancyhead[C]{This paper will be published in 56th IEEE/ACM International Symposium on Microarchitecture (MICRO 2023)} 
  \fancyfoot[C]{}
}

\begin{document}

\title{Dadu-RBD: Robot Rigid Body Dynamics Accelerator with Multifunctional Pipelines} 
\author{Yuxin Yang, Xiaoming Chen*, Yinhe Han*\\
Institute of Computing Technology, Chinese Academy of Sciences\\
Email: yangyuxin18g@ict.ac.cn, chenxiaoming@ict.ac.cn, yinhes@ict.ac.cn\\
*Corresponding authors
\thanks{This work was supported by National Key R\&D Program of China (No. 2018YFA0701500), by National Natural Science Foundation of China (Nos. 62122076, 61834006, and 62025404), by Key Research Program of Frontier Sciences, CAS (No. ZDBS-LY-JSC012), and by Strategic Priority Research Program of CAS (No. XDB44000000).}}

\maketitle

\thispagestyle{firstpage}
\begin{abstract}

  Rigid body dynamics is a core technology in the robotics field. 
  In trajectory optimization and model predictive control algorithms, there are usually a large number of rigid body dynamics computing tasks. 
  Using CPUs to process these tasks consumes a lot of time, which will affect the real-time performance of robots. 
  To this end, we propose a multifunctional robot rigid body dynamics accelerator, named Dadu-RBD, to address the performance bottleneck. 
  By analyzing different functions commonly used in robot dynamics calculations, we summarize their relationships and characteristics, then optimize them according to the hardware. 
  Based on this, Dadu-RBD can fully reuse common hardware modules when processing different computing tasks. 
  By dynamically switching the dataflow path, Dadu-RBD can accelerate various dynamics functions without reconfiguring the hardware. 
  We design the Round-Trip Pipeline and Structure-Adaptive Pipelines for Dadu-RBD, which can greatly improve the throughput of the accelerator. 
  Robots with different structures and parameters can be optimized specifically. 
  Compared with the state-of-the-art CPU, GPU dynamics libraries and FPGA accelerator, Dadu-RBD can significantly improve the performance.
  
\end{abstract}





\section{Introduction} 

Robotics is in a stage of rapid development. 
As demands continue to increase, robots need to face more complex and dynamic environments. 
Various robot structures also require different methods to ensure the balance of the robot itself and the ability to interact with the environment. 
Planning and control algorithms are the key technologies to ensure the robot's ability to move. 

\begin{figure}[!tb]
  \centering
  \includegraphics[width=0.98\columnwidth]{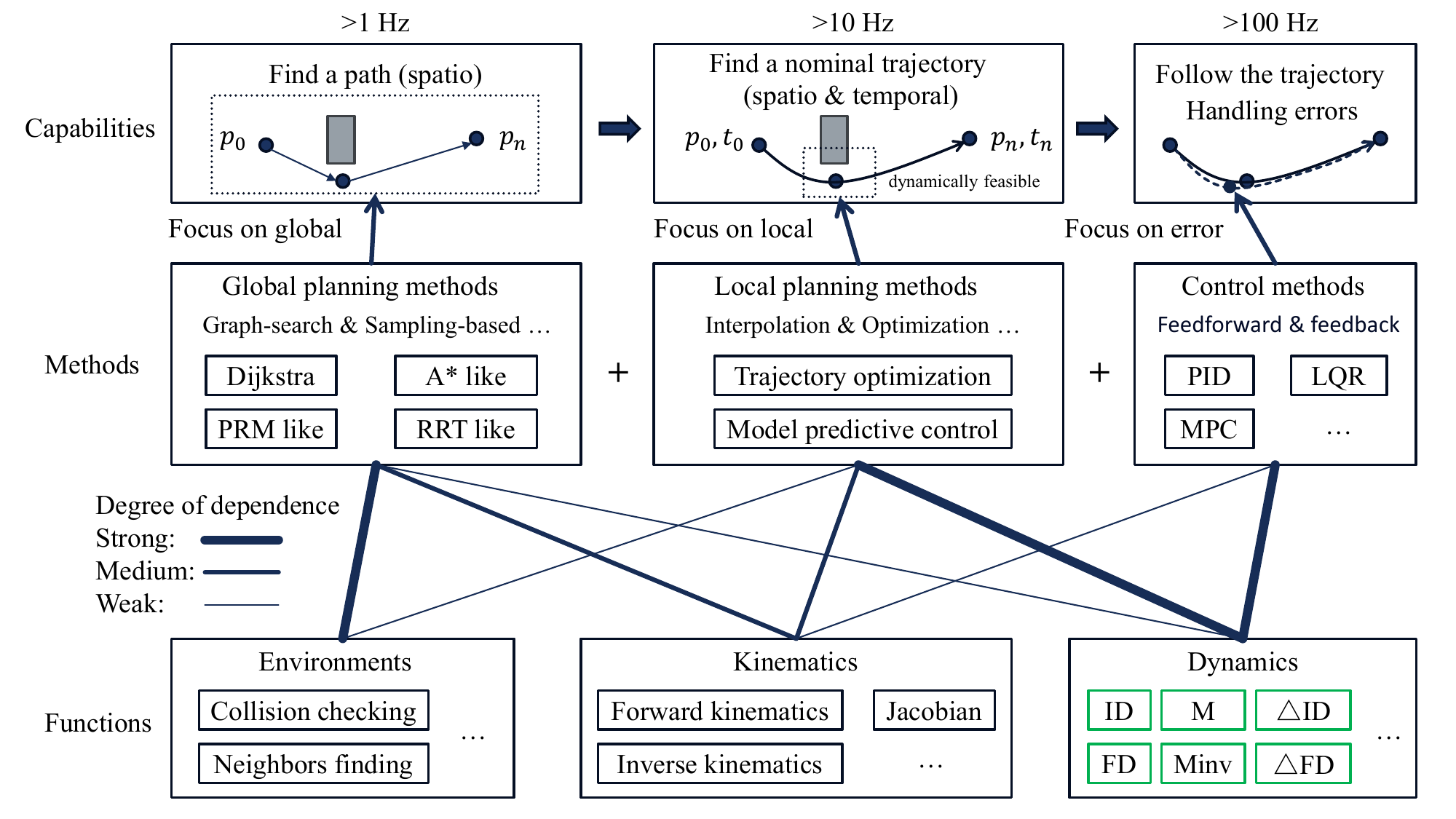}
  \caption{Planning and control framework in robotics.}
  \label{Fig.robot_planning}
\end{figure}

Traditionally, the planning and control framework is shown in Fig.~\ref{Fig.robot_planning}. 
The planning algorithms are divided into two parts: global planning methods and local planning methods. 
The global planning methods mainly focus on the overall spatial path, while the local planning methods mainly focus on the spatio-temporal trajectory in a small area. 
They form the front-end and back-end of the planning algorithms. The front-end can provide a reliable initial value for the optimization algorithm in the back-end. 

There are many different global planning methods, such as A*, PRM(*), RRT(*), and a large number of improved methods derived from them. 
The main purpose of these algorithms is to find a feasible path without obstacles in the space. 
Therefore they strongly rely on environment-related basic functions such as collision detection and nearest neighbor search. 
With the continuous improvement of robot performance requirements, hardware acceleration work related to global planning methods continues to emerge, such as \cite{cd-micro16,DaDu-Series,Yang_Lian_Chen_Han_2020,bakhshalipour2022racod, Jia_Yang_Hsiao_Cruz_Brooks_Wei_Reddi_2022, li2020high}. 

Trajectory optimization (TO) and model predictive control (MPC) \cite{6386025, FastMPC, 8593840} are two important types of local planning methods. 
The MPC algorithm can also be used as a control method. 
The difference is that the MPC algorithm in planning method will output a nominal trajectory for later following, while the MPC algorithm in control method will output the underlying execution instructions directly, which requires higher performance (>100Hz) \cite{9560742, 9783060}. 
The TO and MPC algorithms contain a large number of rigid body dynamics function calls, which will seriously affect the real-time performance of robots \cite{FastMPC, Robomorphic}. 
Depending on the complexity of the robot's structure, the time of the rigid body dynamics computing can account for 30\%-90\% of the total algorithm running time \cite{FastMPC, PinocchioRef, Plancher_Kuindersma_2020}. 
To meet the real-time requirements, some existing works accelerate the TO and MPC algorithms through multi-threaded CPUs \cite{MPC-ICHR-2017, 7525066, Hereid_Harib_Hartley_Gong_Grizzle_2019}. 
There are also some works exploiting the parallelism of GPUs to accelerate these algorithms \cite{arora2009fast, Antony_Grant_2017, 10.1145/3309486.3340246, Guhathakurta_Rastgar_Sharma_Krishna_Singh_2022, Heinrich_Zoufahl_Rojas_2015}. 

In fact, most algorithms in robot dynamics has forward and backward propagation, similar to the training algorithm of neural networks. 
It is a cache-unfriendly calculation, and require a lot of memory access \cite{PinocchioRef,Featherstone_2008}. 
When using CPU multithreading for acceleration, memory bottlenecks will be encountered. 
GPU also has this problem. And the single-task latency of the GPU is relatively large, which is not a good choice for some tasks. 
In addition, high-performance CPUs or GPUs generally have high power consumption, so they are not very suitable for battery-dependent mobile robots. 

\begin{figure}[!t]
  \centering
  \includegraphics[width=0.99\columnwidth]{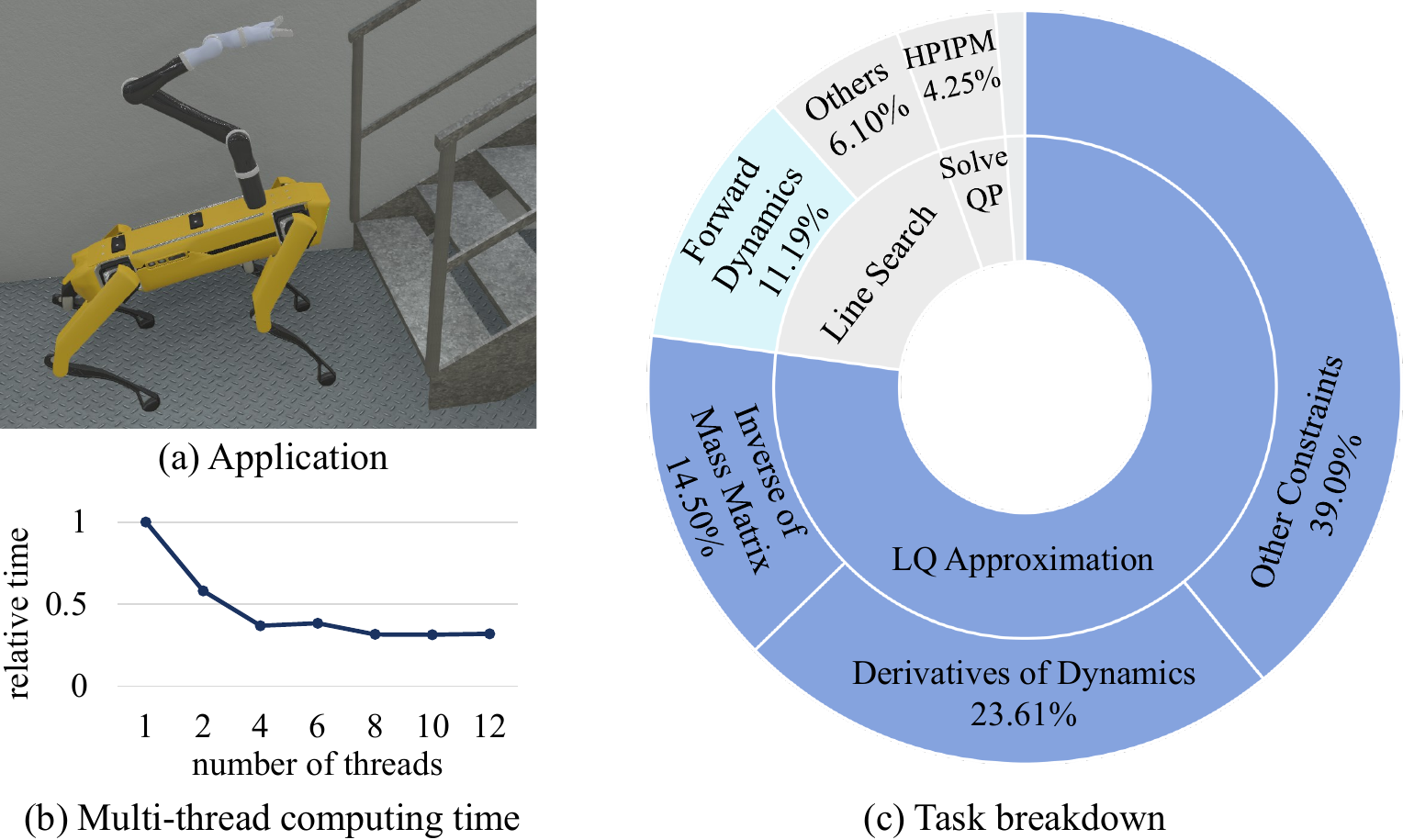}
  \caption{Robot application and performance analysis.}
  \label{Fig.robot_example}
\end{figure}

To verify the above feature, we build a robot application example (Fig.~\ref{Fig.robot_example}a) and test it in the robot simulator Webots \cite{Webots} with the control framework OCS2~\cite{OCS2}. 
As shown in Fig.~\ref{Fig.robot_example}b, when the number of threads increases to a certain level, the performance is no longer significantly improved. 
On the other hand, from Fig.~\ref{Fig.robot_example}c, we can observe that the proportion of parallelizable parts (LQ Approximation, dark blue) is large and contains various types of tasks. 
This implies that the entire task has the potential to be further accelerated in parallel.

For energy-efficient real-time computing, there are many end-to-end domain-specific hardware accelerators designed for the MPC algorithms, such as \cite{6915218, 6852127, Peyrl_Ferreau_Kouzoupis_2015, 2464171, KHUSAINOV2018105, MCINERNEY2018381, RoboX}. 
But they can only deal with linear models or simple robot structures, and cannot handle dynamics models for robots with high degrees-of-freedom (DOF). 
Robomorphic \cite{Robomorphic} tries to use FPGA/ASIC to accelerate rigid body dynamics derivatives. 
But it only supports this single function (forward dynamics derivatives) in the dynamics calculations, and the throughput is not high. 
In addition, it also require the cooperation of the CPU. 
This will bring a lot of communication overhead and extra computational overhead to the CPU. 

In response to the above problems, we design a multifunctional robot dynamics accelerator, Dadu-RBD. 
It is also a general rigid body dynamics accelerator design framework that can be applied to a wide variety of robots. 
Dadu-RBD needs to solve three design challenges. 
The first is how to support multifunction related to rigid body dynamics. 
The second is how to achieve high performance under memory bottlenecks. 
The third is how to reduce the resource usage when dealing with different robot models. 

For the first challenge, the most direct method is to design a separate module for each function, but such a waste of resources is unacceptable. 
We summarize the relationships of commonly used dynamics algorithms, and analyzes the characteristics of them. 
On this basis, we carry out hardware-specific optimizations for these algorithms with the goal of minimizing resource usage while ensuring that there is no major loss in performance. 
At the same time, we designed a global data-driven state machine for the entire Dadu-RBD architecture, so that Dadu-RBD can dynamically switch the dataflow and output results according to the function type, without the need for external instruction sets to control. 
This allows Dadu-RBD to perform calculations independently of the CPU, which can greatly reduce the communication overhead and further improve the performance of the entire system. 

To deal with the second challenge, we design the Round-Trip Pipeline (RTP). 
It is a medium-grained pipeline at joints level, which has deep pipeline stages that support massive parallelism. 
Previous work \cite{Robomorphic} lacked this pipeline depth. 
They use two big cores to process the forward propagation and the backward propagation respectively. 
Each core needs to have the general ability to handle all joints, so it cannot be completely optimized for sparsity and takes up more resources. 
Instead we use as few resources as possible in each submodule for a joint. 
This strategy can generate a dataflow pattern similar to a systolic array, allowing data transmission and computing time to overlap each other, thereby greatly improving throughput while keeping resource consumption low at the expense of a small increase in latency. 
RTP can naturally use the FIFO buffer to store the temporary data required by the pipeline, thereby avoiding the large amount of data exchange between the cache and the main memory in the traditional architecture, and reducing the memory access bandwidth pressure. 

Whether the third challenge can be solved is the key to the practicality of the whole architecture. 
We design the Structure-Adaptive Pipelines (SAPs) for Dadu-RBD.
It uses the medium-grained pipeline strategy mentioned above, and can be optimized according to the structure of the robot itself. 
By utilizing the software and hardware co-optimization, we implement resource reuse and calculation sparsity according to certain strategies, which can greatly reduce resource usage while ensuring throughput. 

We evaluate our architecture using the same FPGA chip as that used in Robomorphic~\cite{Robomorphic}. 
Compared with the existing state-of-the-art CPU\cite{RBDAcc}, GPU\cite{RBDAcc} and FPGA accelerators\cite{RBDAcc, Robomorphic}, Dadu-RBD can achieve 10.3$\times$, 3.4$\times$ and 6.3$\times$ higher throughput, respectively, in the derivatives of dynamics calculations.
Compared with the CPU dynamics library Pinocchio~\cite{PinocchioRef} and GPU dynamics library GRiD~\cite{GRiD_2022}, the performance of Dadu-RBD has also been significantly improved. 

Our main contributions are summarized as follows: 
\begin{itemize} 
  \setlength\itemsep{0em} 
  \item Dadu-RBD implements a high-performance, multifunctional domain-specific hardware accelerator for robot rigid body dynamics. 
  \item We summarize the relationships and characteristics of commonly used dynamics algorithms, and propose hardware-specific optimizations for these algorithms, including a new mass matrix generation/inversion algorithm. 
  \item RTP is proposed to improve the throughput. We implemented all necessary basic dynamics functions using RTP, and carried out corresponding software-hardware co-optimization. 
  \item SAPs are proposed to optimize the performance and resource consumption. Independent pipeline arrays can further perform asynchronous calculations. 
\end{itemize}

\section{Background} \label{Background}

\begin{figure}[!b]
  \centering
  \includegraphics[width=0.98\columnwidth]{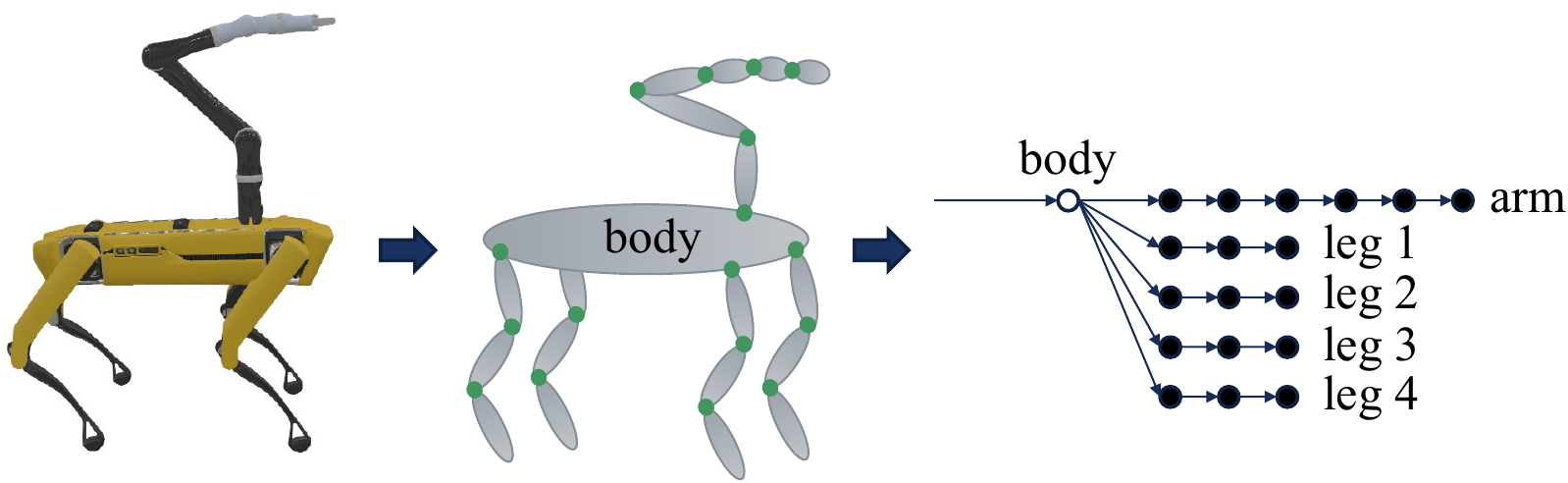}
  \caption{Robot model.}
  \label{Fig.robot_model}
\end{figure}

In order to design a general-purpose robot dynamics accelerator, we need to model the robot in a general format. 
As discussed in \cite{Featherstone_2008}, we can use a topological tree to describe an open-chains robot. 
As the example shown in Fig.~\ref{Fig.robot_model}, the robot has 5 limbs connected to the body. 
Four of them are legs composed of three links, and the remaining one is a robotic arm composed of six links. 
We can assume that the body of the robot is a floating base that connects to a fixed world coordinate system through a virtual 6-DOF joint.
Then each link of the robot can correspond to a joint. 
We can describe each link and joint in the form of matrices.
We assume that the robot has $N_B$ (number of bodies/links) joints and links, with a total of $N$ DOF.
Each link has its own mass and rotational inertia, which can be represented by the symmetric inertia matrix $I_{i} \in \mathbb{R}^{6\times6}$. 
Each joint has a specific type. 
Types of joints include revolute, prismatic, helical, cylindrical, planar, spherical, 3-DOF translation, 6-DOF joint, etc. 
Different types of joints have different motion subspaces $S_{i} \in \mathbb{R}^{6\times N_i}$, where $N_i$ is the DOF of ith joint.  
For the most common joints in robots (revolute and prismatic), $N_i = 1$ and $S_{i}$ are one-hot vectors. 
The definitions of $S_{i}$ for other type of joints can be found in \cite{Featherstone_2008}. 
When the joint state $q_i$ is given, we can calculate the pose relationship between the two links connected by the joint, which can be represented by the transformation matrix ${}^{i}X_{\lambda_i} \in \mathbb{R}^{6\times6}$, where $\lambda_i$ is the parent link's id. 
The transformation matrix ${}^{i}X_{\lambda_i}$ has a unique sparsity. Its top right $3\times3$ elements are always 0. 
The inertia matrix $I_{i}$ also has sparsity according to the joint and link. 
All the above parameters are defined in their respective joint coordinate systems. 

For the robots with the same model, $N_B$, $S_{i}$ and the sparsity of $I_{i}$ are all the same. 
But we may need to calibrate the parameters in $I_{i}$ and ${}^{i}X_{\lambda_i}$ for different robots with the same model. 
For a specific robot, $I_{i}$, $S_{i}$ and parameters in ${}^{i}X_{\lambda_i}$ can be seen as constant.

Most of calculations in robot dynamics are essentially related to the \textbf{equation of motion} for a rigid body system: 
\begin{equation}
  M(q)\ddot q + C(q, \dot q, f^{ext}) = \tau,
  \label{eq-EoM}
\end{equation}
where $q, \dot q, \ddot q, \tau$ are vectors of position, velocity, acceleration and force/torque variables, respectively, 
$M(q)$ is the symmetric positive-definite mass matrix (or joint space inertia matrix) of the robot, 
$C(q, \dot q, f^{ext})$ is the generalized bias forces accounts for the Coriolis and centrifugal forces, gravity, and any other forces ($f^{ext}$) acting on the system other than those in $\tau$. 

Different applications require the calculation of the equation of motion (Eq.~\eqref{eq-EoM}) from different perspectives: 
\begin{itemize}
  \setlength\itemsep{0em}
  \item Find $\tau$ in control algorithms and TO \cite{45152, KATAYAMA20206483, doi:10.1177/0278364912469821, tedrake2019drake, control_toolbox};
  \item Find $\ddot q$ in simulation and MPC \cite{graichen2012real, KATAYAMA20206483, Julia_for_robotics, FastMPC, OCS2};
  \item Find $M$ in optimal control and TO \cite{Crocoddyl, tedrake2019drake, control_toolbox};
  \item Find $M^{-1}$ in kinematics and MPC \cite{Lynch_Park_2017, graichen2012real, Crocoddyl, OCS2};
  \item Find $\partial_u \tau$ in optimal control and TO \cite{Crocoddyl, KATAYAMA20206483, tedrake2019drake, control_toolbox};
  \item Find $\partial_u \ddot q$ in MPC\cite{diehl2006fast, graichen2012real, FastMPC, neunert2018whole, OCS2} ($u=[q;\dot q]$);
\end{itemize}

We organize these calculations into different functions in Table~\ref{table:algo}.
Here var $u$ represents both $q$ and $\dot q$, and $\partial_u \tau$ means $(\frac{\partial \tau}{\partial q}, \frac{\partial \tau}{\partial \dot q})$. 
The definitions of these functions are consistent with those in \cite{Featherstone_2008, PinocchioRef, RBDAcc, Robomorphic}. 
These functions are what Dadu-RBD needs to accelerate.

\begin{table}[!h]
  \centering
  \caption{Rigid body dynamics functions overview.}
  \label{table:algo}
  \begin{tabular}{|l|l|}
    \hline
    \textbf{Function Name} & \textbf{Definition}\\
    \hline
    Inverse Dynamics & $ \tau = \operatorname{ID}(q, \dot q, \ddot q, f^{ext}) $ \\
    \hline
    Forward Dynamics & $ \ddot q = \operatorname{FD}(q, \dot q, \tau, f^{ext}) $ \\
    \hline
    \makecell[l]{Mass Matrix} & $ M = \operatorname{M}(q) $\\
    \hline
    \makecell[l]{Inverse of Mass Matrix} & $ M^{-1} = \operatorname{Minv}(q) $\\
    \hline
    \makecell[l]{Derivatives of ID} & $ \partial_u \tau = \operatorname{\Delta ID}(q, \dot q, \ddot q, f^{ext}) $ \\
    \hline
    \makecell[l]{Derivatives of FD} & $ \partial_u \ddot q = \operatorname{\Delta FD}(q, \dot q, \tau, f^{ext}) $ \\
    \hline
    \makecell[l]{Derivatives of Dynamics} & $ \partial_u \ddot q = \operatorname{\Delta iFD}(q, \dot q, \ddot q, M^{-1}, f^{ext}) $ \\
    \hline
  \end{tabular}
\end{table}

\section{Key Insights} \label{Insights}

\subsection{Relationships in Rigid Body Dynamics} \label{rbd_relationships}

In software algorithms, the different functions of the dynamics are usually independent of each other. 
This is because different algorithms can be optimized for the desired function, and we can choose the one with the best performance. 

However, all functions in Table 1 satisfy the equation of motion (Eq.~\eqref{eq-EoM}), there must be a strong relationship between them. 
Notice that if we fix $q,\dot q, $ and $f^{ext}$ in Eq.~\eqref{eq-EoM}, then the whole equation is actually simplified to the form of $\tau=M\ddot q+C$, where $M, C$ are both constants. 
This means that $\ddot q$ and $\tau$ are two vectors that are linearly related. 
Once we know one of them, we can easily get another. 
This relationship holds for both dynamics and gradients of dynamics \cite{Featherstone_2008, PinocchioRef}. 
That is to say, we have the following relationships in rigid body dynamics: 
\begin{equation}
  \operatorname{FD} = M^{-1}\operatorname{ID},
  \label{eq-ID-FD-relation}
\end{equation}
\begin{equation}
  \operatorname{\Delta FD} = M^{-1}\operatorname{\Delta ID}.
  \label{eq-dID-dFD-relation}
\end{equation}

The existence of this relationship is very friendly to hardware design, allowing our multifunctional accelerators to reuse the computing resources. 
The function $\operatorname{ID}$ can be calculated using the Recursive Newton-Euler Algorithm (RNEA) \cite{Featherstone_2008}, and the function $\operatorname{\Delta ID}$ can also be analytically calculated by $\operatorname{\Delta RNEA}$ \cite{PinocchioRef}. 
In software, the function $\operatorname{FD}$ is usually calculated using the efficient Articulated Body Algorithm (ABA) \cite{Featherstone_2008}. 
But its gradient calculation is relatively complicated, so we can use the relationship Eq.~\ref{eq-dID-dFD-relation} to get $\operatorname{\Delta FD}$ from the function $\operatorname{\Delta ID}$. 
For further resource sharing, we can completely omit the instantiation of the ABA algorithm, and also use the relationship Eq.~\ref{eq-ID-FD-relation} to implement the $\operatorname{FD}$ function. 

There is also a strong relationship between the calculation of the mass matrix and its inverse. 
There are many ways to compute the inverse of the mass matrix. 
We can first use the Composite Rigid Body Algorithm (CRBA) \cite{Featherstone_2008} to calculate the mass matrix, and then use a certain decomposition algorithm to calculate the inverse of the mass matrix. 
Because the mass matrix is a symmetric positive definite matrix, it can be decomposed by Cholesky factorization or $LDL^T$ factorization, which can be written as $M = LDL^T$. 
After the decomposition, we can get $M^{-1}$ efficiently \cite{6710599}. 

In many cases, the inverse of the mass matrix is calculated to aid in the calculation of forward dynamics. 
If we just want to calculate the forward dynamics, we do not even need to know what $M^{-1}$ is. 
We can transform Eq.~\eqref{eq-EoM} into $L^{-T}D^{-1}L^{-1} \ddot q = (\tau - C)$, 
then we can solve $\ddot q$ by two back-substitution procedures. 
But this method has shortcomings. 
We use the CRBA to first calculate the mass matrix, and then calculate the inverse of the matrix or solve the linear equation system through matrix decomposition. 
The two serial steps will introduce long latency when accelerated in parallel using dedicated hardware. 

In fact, we can do part of the matrix decomposition and back-substitution procedures in advance while computing the $M$ matrix, especially the reciprocal operation. 
This can overlap the computational latency of the two parts, thereby reducing the latency of the overall computation. 
This also allows the reuse of resources for computing the mass matrix and its inverse. 
More detailes will be discussed in a Section~\ref{rtp-MMinv}.

\subsection{Characteristics of Rigid Body Dynamics}

In order to describe the characteristics of the dynamic algorithm, we first introduce the most basic dynamics algorithm RNEA, as shown in Algorithm~\ref{alg:RNEA}. 
We modified it a little in detail to meet the needs of the hardware design, which will be mentioned in Section~\ref{rtp-id}. 
It can be observed that the RNEA algorithm has a serial process of forward propagation and backward propagation. 
The intermediate results obtained in the forward propagation need to be used in the backward propagation. 

\begin{algorithm}[!t]
  \small
  \caption{$\operatorname{RNEA}$}
  \label{alg:RNEA}
  \begin{algorithmic}[1]
    \REQUIRE $q, sinq, cosq, \dot q, \ddot q, f^{ext}$
    \ENSURE $\tau, [v, a, f]$
    \STATE $v_0 = 0, a_0 = gravity$
    \FOR{$i = 1:N_B$} 
      \STATE ${}^iX_{\lambda_i} = X_i(q_i, sinq_i, cosq_i)$
      \STATE $v_i = {}^{i}X_{\lambda_i}v_{\lambda_i} + S_i \dot{q_i}$
      \STATE $a_i = {}^{i}X_{\lambda_i}a_{\lambda_i} + S_i \ddot{q_i} + v_i \times S_i \dot{q_i}$
      \STATE $f_i = I_i a_i + v_i \times^* I_i v_i - f^{ext}_i$
    \ENDFOR
    \FOR{$i = N_B:1$} 
      \STATE ${}^{\lambda_i}X_i^* = X_i^T(q_i, sinq_i, cosq_i)$
      \STATE $\tau_i = S_i^T f_i$
      \STATE $f_{\lambda_i} += {}^{\lambda_i}X_i^* f_i$
    \ENDFOR
  \end{algorithmic}
\end{algorithm}

\begin{figure}[!t]
  \centering
  \subfloat[Computing feature]
  {\includegraphics[width=.45\columnwidth]{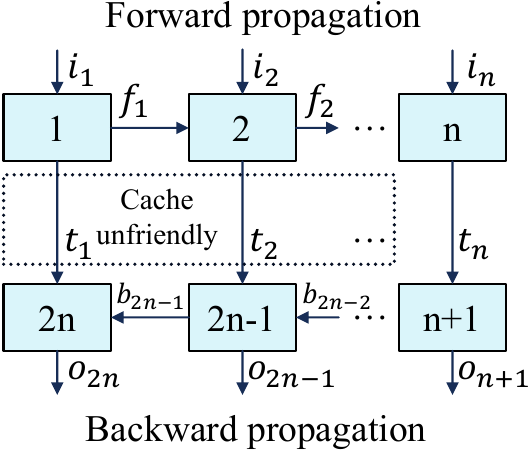}
      \label{Fig.insight_compare_feature}
  }\hspace{0pt}
  \subfloat[Behavior of CPU/Cache]
  {\includegraphics[width=.45\columnwidth]{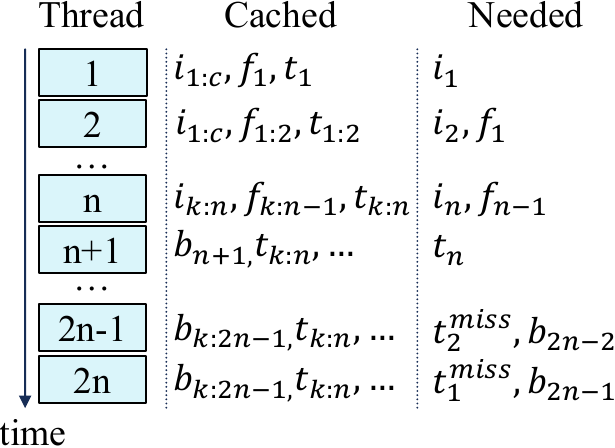}
      \label{Fig.insight_compare_cpu}
  }\\
  \subfloat[Behavior of Robomorphic]
  { \includegraphics[width=.45\columnwidth]{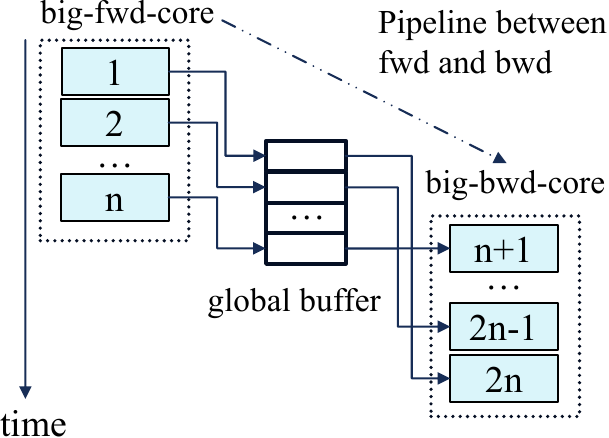}
      \label{Fig.insight_compare_robom}
  }\hspace{0pt}
  \subfloat[Behavior of RTP]
  {\includegraphics[width=.45\columnwidth]{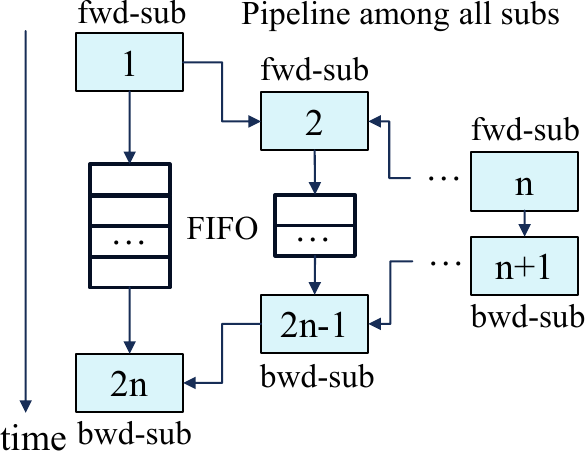}
      \label{Fig.insight_compare_rbdcore}
  }
  \caption{Analyze the behavior of different hardware architectures for the same computing feature.}
  \label{Fig.insight_compare}
\end{figure}

This feature also exists for other dynamic algorithms, which can be physically explained as a chain-like dependency between robot joints. 
We abstract this computing feature, as shown in Fig.~\ref{Fig.insight_compare_feature}. 
The data transfer between forward pass and back pass is hardware unfriendly. The process of forward and backward propagation leads to poor temporal locality of intermediate data. 
For the modern CPU/Cache architecture, these intermediate results are easily missed in the cache, thus indirectly increasing the memory access pressure, as shown in Fig.~\ref{Fig.insight_compare_cpu}.

Actively controlling the on-chip buffer can alleviate this problem, so using hardware such as FPGAs for acceleration is an effective way. 
Robomorphic~\cite{Robomorphic} focuses on optimizing the latency of calculations, so it makes high-performance computing cores for forward propagation and back propagation respectively, as shown in Fig.~\ref{Fig.insight_compare_robom}. 
It uses a global on-chip buffer to store intermediate variables. 
Between the forward core and the backward core, pipelines can be used to improve throughput to a certain extent. 
These two ``big'' computing cores are applicable to all joints, so much more resources are required. 

In contrast, Dadu-RBD focuses on optimizing the computation throughput. 
We propose Round-Trip Pipeline to achieve massively pipelined parallelism, as shown in Fig.~\ref{Fig.insight_compare_rbdcore}. 
The FIFO buffer between the forward and backward propagation acts as a bypass buffer for the pipeline, keeping the entire pipeline running efficiently and reducing the amount of off-chip memory access. 
Here, each submodule only cares about one joint, and targeted sparsity optimization can be performed on it. 
At the same time, it is noted that the computing load of different submodules will vary with the depth of the joint, so they can have different optimization strategies. 
Submodules with heavy loads need to ensure computing latency to maintain the frequency of the overall pipeline; 
Submodules with light loads can fully reuse computing resources without affecting the overall pipeline frequency.

\subsection{Adapte to the robot structure}

\begin{figure}[!tb]
  \centering
  \includegraphics[width=0.8\columnwidth]{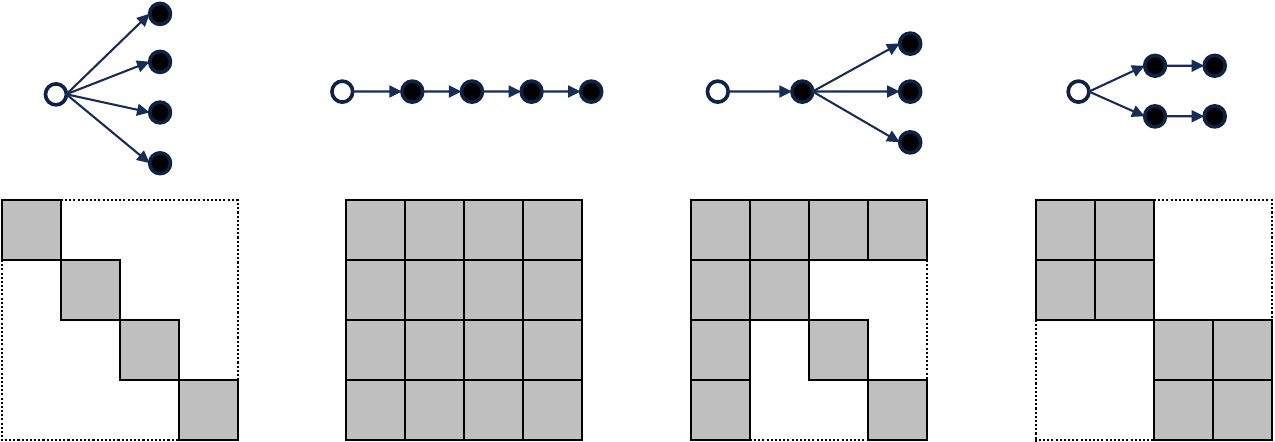}
  \caption{Branch induced sparsity}
  \label{Fig.branch_induced_sparsity}
\end{figure}

Robot structures in the real world are diverse, and Dadu-RBD needs to be guided by a general method as a robot dynamics accelerator. 
We note that the structure of the robot can have a huge impact on the sparsity of the robot dynamics parameter matrix \cite{Featherstone_2008} (see Fig.~\ref{Fig.branch_induced_sparsity}). 
In Algorithm~\ref{alg:RNEA}, we can also see that the robot structure also has an important impact on the topology of the robot dynamics calculation. 
These effects are all due to the branching structure of the robot. 
The intuitive understanding is that the dynamics of a certain joint of the robot will only be propagated to the branchs it can affect. 
After introducing the forward and backward propagation process of the dynamics algorithm, the sparsity complexity of the matrix will be further increased. 
These sparsity and computational complexity changes due to robot topology are important issues that we must consider.

We address this issue by proposing Structure-Adaptive Pipelines, which optimize the accelerator architecture according to the robot architecture. 
SAPs contains a set of software and hardware collaborative optimization methods. 
It will provide effective hardware support according to the branch characteristics of the robot, including time division multiplexing, sparsity optimization, etc. 
The topology of the robot will also be properly rotated according to the hardware characteristics, so as to reduce the overall resource consumption. 

\section{Round-Trip Pipeline Design}

\subsection{Inverse Dynamics and Its Derivatives} \label{rtp-id}

\begin{figure}[!b]
  \centering
  \includegraphics[width=0.98\columnwidth]{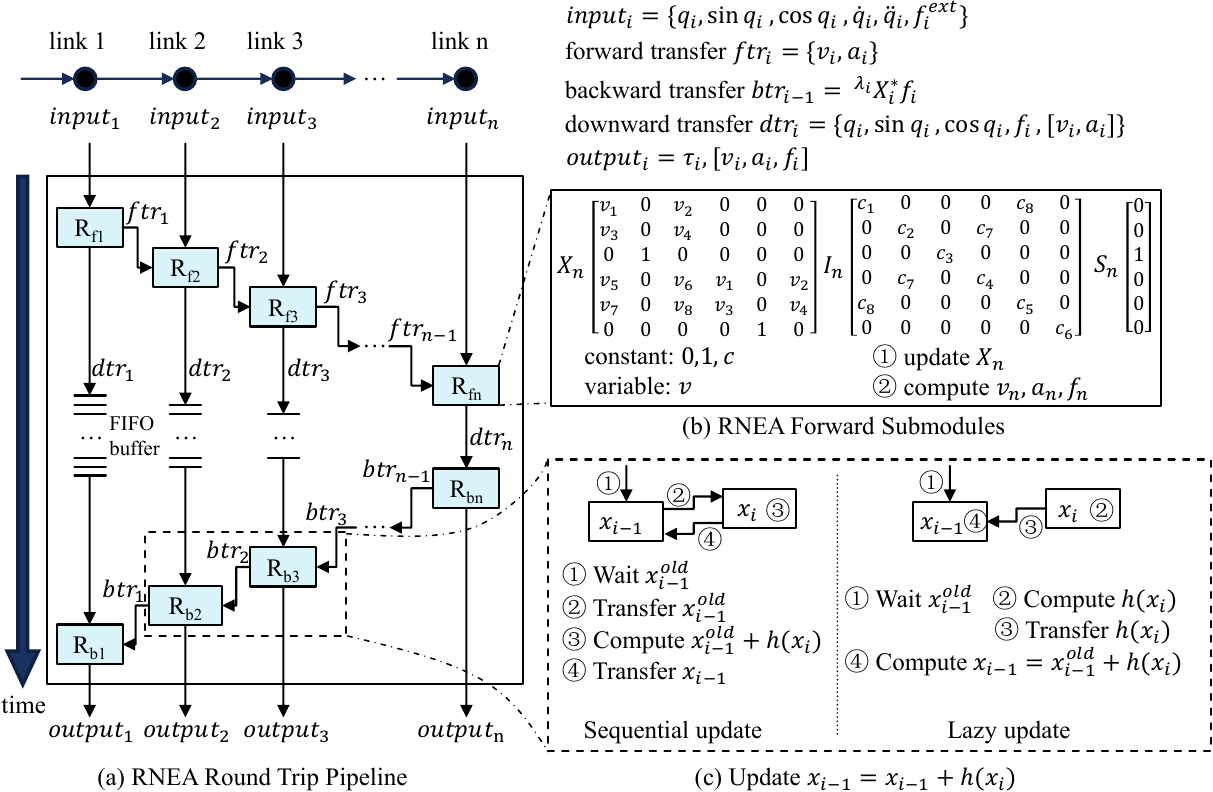}
  \caption{RTP Design for $\operatorname{RNEA}$}
  \label{Fig.hardware_rnea}
\end{figure}

\begin{figure}[!b]
  \centering
  \includegraphics[width=0.98\columnwidth]{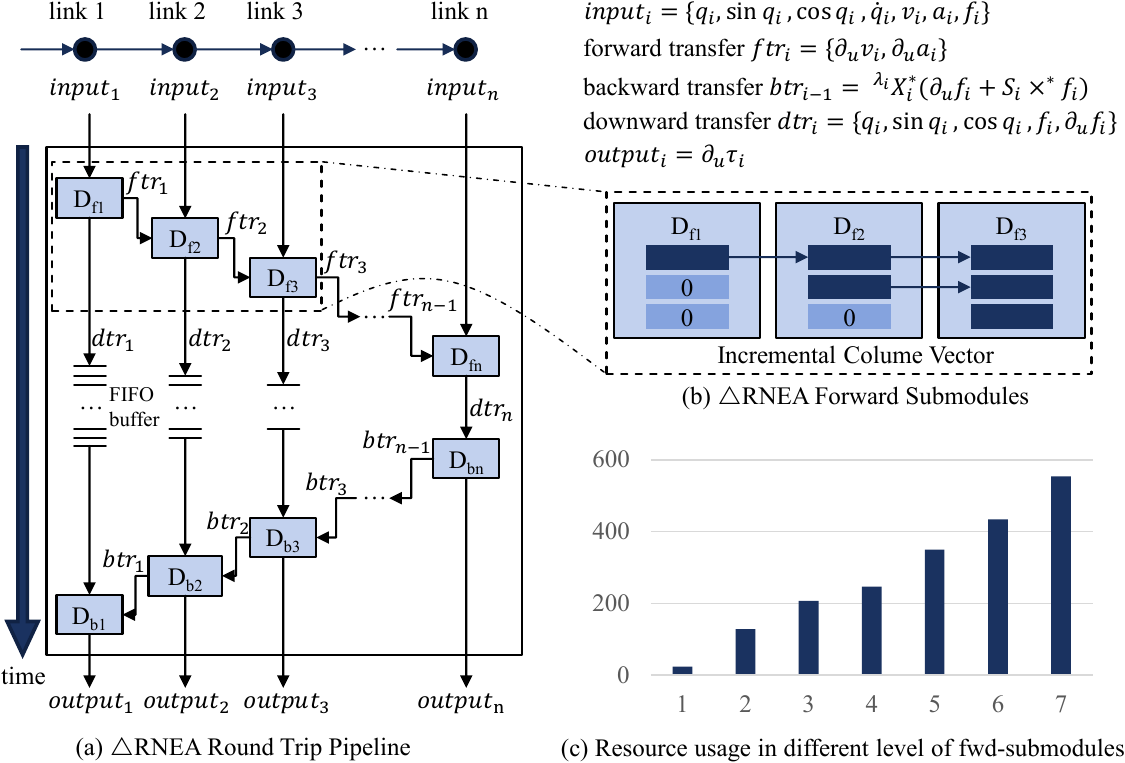}
  \caption{RTP Design for $\operatorname{\Delta RNEA}$}
  \label{Fig.hardware_drnea}
\end{figure}

We design the $\operatorname{RNEA}$ RTP and the $\operatorname{\Delta RNEA}$ RTP, as shown in Fig.~\ref{Fig.hardware_rnea} and Fig.~\ref{Fig.hardware_drnea}. 

In the $\operatorname{RNEA}$ RTP, each iteration of the loop in the Algorithm.~\ref{alg:RNEA} is mapped to a submodule. 
For example, for a robot arm with $N_B=n$ links, the $\operatorname{RNEA}$ RTP has $2n$ submodules, namely $R_{fi}$ and $R_{bi}$, where $i \in [1, n]$; 
These submodules are independent of each other and transmit data through FIFO streams. 
In this way, each submodule can just handle the calculation of the corresponding joint and link. 
For the submodule $R_{fi}$, it needs to update $X_i$, calculate $v_i,a_i,f_i$ according to the Algorithm~\ref{alg:RNEA}, then transfer $ftr_i$ forward to $R_{fi+1}$ and $dtr_i$ downward to $R_{bi}$. 
For the submodule $R_{bi}$, it needs to update $X_i$ again, update $f_i$, compute $btr_{i-1}$ and $\tau_i$, then transfer $btr_{i-1}$ backward to $R_{bi-1}$ and $\tau_i$ to $output_i$. 

The $\operatorname{\Delta RNEA}$ RTP also has $2n$ submodules, namely $D_{fi}$ and $D_{bi}$, where $i \in [1, n]$. 
It has the same structure as the $\operatorname{RNEA}$ RTP, but the input/output and the calculation content in each submodule are different. 

For each $\operatorname{RNEA}$ or $\operatorname{\Delta RNEA}$ calculation, it needs to go through forward pass and backward pass. 
For the shallower (the subscript is smaller) submodules, intermediate data needs to be cached for a long time. 
Certain bypass buffers are added to our hardware design to avoid pipeline stalls. 
In this way, the multi-stage pipeline can greatly improve the parallelism of calculations, thereby improving throughput. 
Although there are multiple similar matrix multiplications inside the submodule, we did not implement a fine-grained matrix multiplication pipeline unit. 
Doing so will consume a lot of resources and will also make the sparse optimization less efficient. 

We have some methods to further optimize the design. 

\subsubsection{\textbf{Sparsity and Constant Optimization}} 
As shown in Fig.~\ref{Fig.hardware_rnea}b, we can optimize the design according to the fixed sparse characteristics and constants in the robot parameters. 
We assume that the nth joint is the most common revolute joint in robots. 
Then only 12 elements in the $6 \times 6$ matrix $X_n$ are non-constant, and these 12 elements have only 8 different values, all of which are in the form of $c*sinq$ or $c*cosq$. 
The symmetric matrix $I_n$ has only 8 distinct non-zero constants, and the $S_n$ is a one hot vector. 
These features are also mentioned in Section~\ref{Background}, which can greatly reduce the computational cost. 
Therefore, we don't need to design a complete $6 \times 6$ matrix vector computing unit, but only need to consider the non-zero part. 
We can also further optimize the design according to the characteristics of the hardware, such as saving the constants on-chip, or using the look up tables in FPGAs, thereby reducing memory access overhead and improving performance. 

\subsubsection{\textbf{Reupdate Transformation Matrix}} 
In the backward submodules $R_{bi}$ and $D_{bi}$, we reupdate $X_i$. 
Instead of buffer and transfer the calculated $X_i$, we recalculate it. 
As mentioned above, the calculation cost of x is very small, while the transmission cost is relatively greater. 
In matrix $X_i$, only one or two multiplication are required. 
Since the forward submodules are more complex than the backward submodules, we need to minimize the number of interfaces and tasks of the forward module. 
Therefore, we choose to transfer less data downward. 
Only one data $q_i$ is needed for the prismatic joint, and only two data $sinq_i, cosq_i$ are needed for the revolute joint. 

\subsubsection{\textbf{Lazy Update}} 
There are some operations that need to add or subtract variables from other loop bodies, such as Algorithm~\ref{alg:RNEA} line 10, and the corresponding calculation in the $\operatorname{\Delta RNEA}$ algorithm. 
If we update the data sequentially as described by the original algorithm, the loopback dependency between loop bodies will destroy the dataflow of the pipeline. 
Actually we do not need to read them first and then write them back. 
We can just pass the addend or subtraction to the corresponding submodule, and wait for the next cycle to calculate. 
We call this approach lazy update, as shown in Fig.~\ref{Fig.hardware_drnea}c. 

\subsubsection{\textbf{Incremental Calculation}} \label{Incremental-Calculation}
In the $\operatorname{\Delta RNEA}$ RTP, the matrices have different sparsity characteristic related to the depth of iterations. 
In fact, the number of their useful columns is proportional to the iteration depth, as shown in Fig.~\ref{Fig.hardware_drnea}b. 
We can further utilize column vectors (columns of the matrix) to only pass and calculate useful columns, while each submodule only needs to incrementally initialize newly added columns. 
This feature will cause deeper submodules to have higher computation load, for which we need to allocate more computing resources to them (Fig.~\ref{Fig.hardware_drnea}c). 
Due to the increase in calculations, deeper sub-modules will inevitably become the performance bottleneck of the entire RTP, so we can adopt a more aggressive resource reuse strategy for shallower modules to reduce overall resource consumption.

\subsection{Mass Matrix and Its Inverse} \label{rtp-MMinv}

\begin{algorithm}[!t]
  \small
  \caption{$\operatorname{MMinvGen}$}
  \label{alg:MMinv}
  \begin{algorithmic}[1]
    \REQUIRE $q, sinq, cosq, out_M, out_{Minv}$
    \ENSURE $M$ or $M_{inv}$

    \FOR{$i = N_B:1$} 
      \STATE ${}^iX_{\lambda_i} = X_i(q_i, sinq_i, cosq_i); {}^{\lambda_i}X_i^* = {}^iX_{\lambda_i}^T$
      \STATE $I_i^A \pluseq I_i; U_i = I_i^A S_i; D_i = S_i^T U_i$
      \IF{$out_{Minv}$}
        \STATE $M_{inv}[i,i] = D_i^{-1}$
        \STATE $M_{inv}[i,tree_e(i)] = -D_i^{-T} S_i^T F_i[:,tree_e(i)]$
      \ENDIF
      \IF{$out_M$}
        \STATE $M[i,i] = D_i$
        \STATE $M[i,tree_e(i)] = F_i^T[:,tree_e(i)] S_{tree_e(i)}$
      \ENDIF
      \IF{$\lambda_i \neq 0$} 
        \IF{$out_{Minv}$}
          \STATE $F_i[:,tree(i)] \pluseq U_iM_{inv}[i,tree(i)]$
          \STATE $I_i^A \minuseq U_i D_i^{-1} U_i^T$
        \ENDIF
        \IF{$out_M$}
          \STATE $F_i[:,i] = U_i$
        \ENDIF
        \STATE $F_{\lambda_i}[:,tree(i)] \pluseq {}^{\lambda_i}X_i^* F_i[:,tree(i)]$
        \STATE $I_{\lambda_i}^A \pluseq {}^{\lambda_i}X_i^* I_i^A {}^iX_{\lambda_i}$
      \ENDIF
    \ENDFOR
    \IF{$out_{Minv}$}
      \FOR{$i = 1:N_B$}
        \IF{$\lambda_i \neq 0$} 
          \STATE $M_{inv}[i,\mathrel{{i}{:}}] \minuseq D_i^{-1} U_i^T {}^iX_{\lambda_i} P_{\lambda_i}[:,\mathrel{{i}{:}}]$
        \ENDIF
        \STATE $P_i[:,\mathrel{{i}{:}}] = S_i M_{inv}[i,\mathrel{{i}{:}}]$
        \IF{$\lambda_i \neq 0$} 
          \STATE $P_i[:,\mathrel{{i}{:}}] \pluseq {}^iX_{\lambda_i} P_{\lambda_i}[:,\mathrel{{i}{:}}] $
        \ENDIF
      \ENDFOR
    \ENDIF
  \end{algorithmic}
\end{algorithm}
%

\begin{figure}[!t]
  \centering
  \includegraphics[width=0.98\columnwidth]{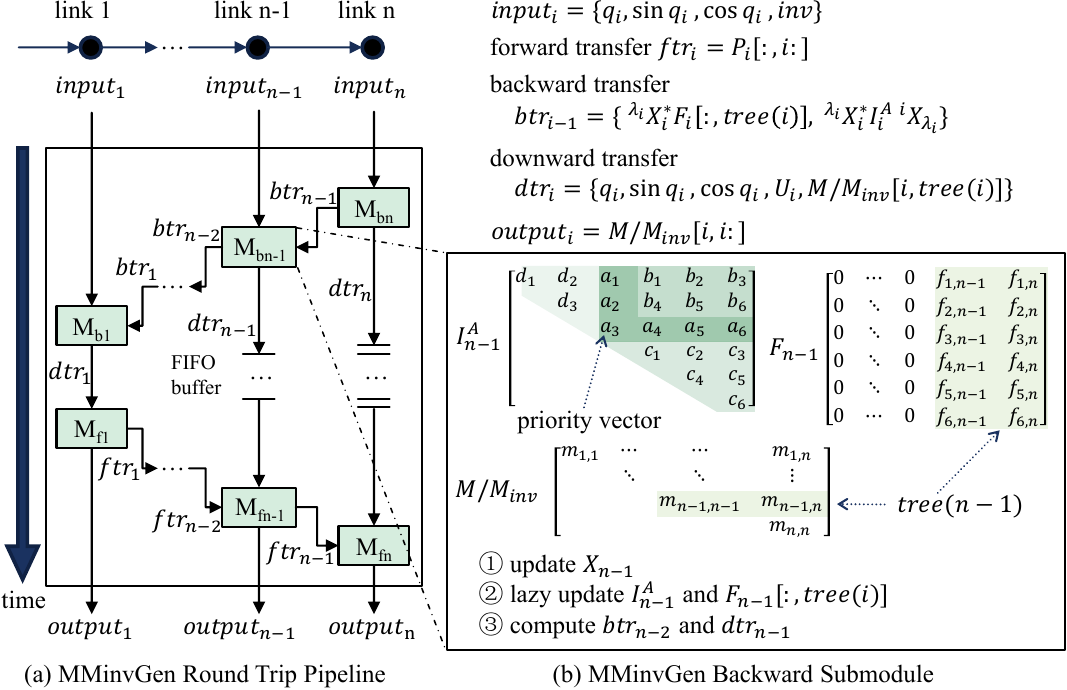}
  \caption{RTP Design for $\operatorname{MMinvGen}$}
  \label{Fig.hardware_mminv}
\end{figure}

We propose $\operatorname{MMinvGen}$ (Algorithm~\ref{alg:MMinv}) to generate the mass matrix or the inverse of the mass matrix. 
It combines the CRBA algorithm \cite{Featherstone_2008} and a simplified ABA algorithm \cite{carpentier2018analytical}. 
We analyze the differences and similarities of these two algorithms, and further simplify and unify them through software-hardware co-optimization. 
Compared with the original algorithms, we avoid a whole forward loop, which can greatly reduce hardware resource consumption. 
Compared with the CRBA algorithm, our algorithm greatly improves the parallelism of computation. 
We can choose whether to generate the mass matrix through the $out_M$ parameter, and whether to do matrix factorization and inversion then generate the inverse of the mass matrix through the $out_{Minv}$ parameter. 
The intermediate variables $I_i^A$ and $F_i$ are initialized to all-zero matrices. 
The notation $tree(i)$ represents the set of id of all nodes contained in the subtree of node i, and $tree_e(i) = tree(i) \backslash i$.

Based on Algorithm~\ref{alg:MMinv}, we design the $\operatorname{MMinvGen}$ RTP, as shown in Fig.~\ref{Fig.hardware_mminv}. 
The overall structure of the $\operatorname{MMinvGen}$ RTP is basically similar to that of the RNEA RTP, but the direction of the dataflow is reversed. 
It uses all the optimization methods used in $\operatorname{RNEA}$ RTP and $\operatorname{\Delta RNEA}$ RTP, 
such as the Incremental Calculation and the Lazy Update for the matrix $F_i$, and the Sparsity Optimization for the symmetric matrices $M, M^{-1}$ and $I^A_i$, as show in Fig.~\ref{Fig.hardware_mminv}b.

In addition, it has the following optimization methods:
\subsubsection{\textbf{Priority Vector}}
The symmetric matrix $I^A_i$ needs to be updated lazily in each submodule, yet its computation is in the critical path of the entire pipeline. 
We no longer divide it into column vectors in the previous way, but divide it into four vectors according to specific priorities, calculate and pass the vectors in the critical path first. 
Taking the revolute joint as an example, the calculation of the third column element of the symmetric matrix $I^A_i$ is on the critical path, so we set it as the priority vector.
The division method is shown in Fig.~\ref{Fig.hardware_mminv}b.

\subsubsection{\textbf{Fixed-point vs Floating-point Reciprocal}}
Compared with floating-point numbers, fixed-point addition, subtraction and multiplication are very resource-efficient and fast. 
However, there are reciprocal operations in Algorithm~\ref{alg:MMinv} line 5. 
And fixed-point division is very slow. 
Therefore, we need to convert fixed-point numbers into floating-point numbers first, and then use the characteristics of floating-point numbers to quickly find the reciprocal \cite{Istoan_Pasca_2015}. 
After getting the result, it is converted back to a fixed-point number to participate in subsequent operations. 
This can greatly improve performance.

\section{Architecture of Dadu-RBD}

The hardware in the previous section is designed for a serial robotic arm with a fixed base, but the actual robot has a more complex structure and does not have a fixed base. 
Dadu-RBD needs to take this into account. 
Based on this, the versatility of Dadu-RBD needs to have two aspects, supporting multiple functions and supporting multiple robots. 
It should be noted that Dadu-RBD's support for multi-robots is not dynamic, but needs to be reconfigured. For a specific model of robot, only once initial configuration is required. 
In this section, we discuss how Dadu-RBD optimizes performance and resource usage while implementing a general robot rigid body dynamics accelerator.

\subsection{Multifunction Optimization}

\begin{figure}[!tb]
  \centering
  \includegraphics[width=0.98\columnwidth]{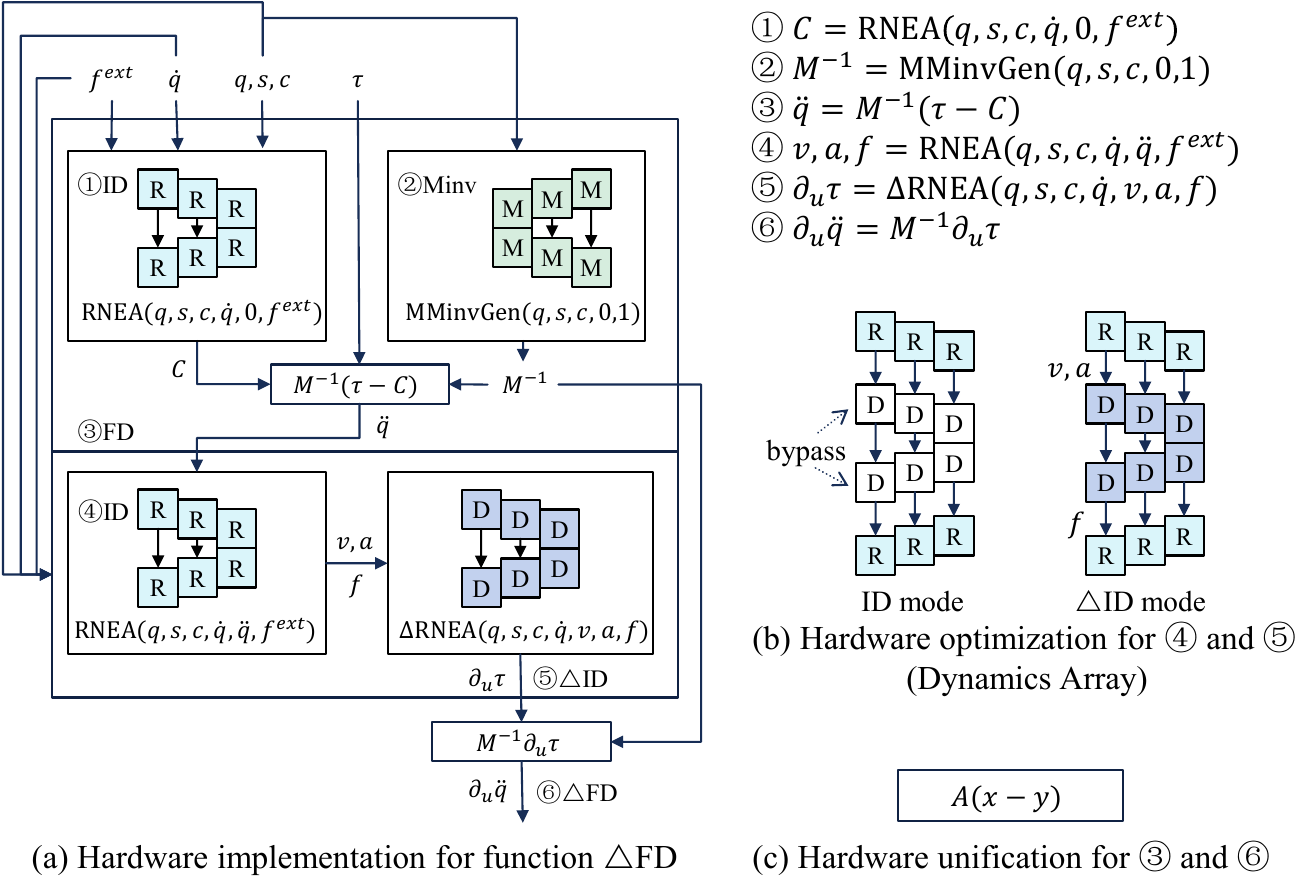}
  \caption{Hardware optimization for multifunction.}
  \label{Fig.hardware_optimization}
\end{figure}

As mentioned in Section~\ref{rbd_relationships}, thanks to the close relationships amoung the robot rigid body dynamics calculations, we can fully reuse hardware resources for different robot rigid body dynamics functions. 

It is worth noting that in order to completely calculate the result of $\operatorname{\Delta FD}$, we must go through 6 steps (\ding{172} to \ding{177}), as shown in Fig.~\ref{Fig.hardware_optimization}a. 
We found that the functions $\operatorname{ID}$, $\operatorname{FD}$, $\operatorname{Minv}$, $\operatorname{\Delta ID}$ and $\operatorname{\Delta FD}$ are actually a subset of these 6 calculation steps. 
Therefore we can implement these functions through a combination of these 6 steps. 
By implementing these calculation steps and using some auxiliary modules, Dadu-RBD can support all the functions in Table~\ref{table:algo}. 
We will cover this in more detail in Section~\ref{dataflow}.

For $\operatorname{RNEA}$ and $\operatorname{\Delta RNEA}$ RTP, the intermediate data $v,a,f$ need to be transferred between their submodules. 
We further optimized the data paths between them, as shown in Fig.~\ref{Fig.hardware_optimization}b. 
We name this submodules' structure the Dynamics Array. 
It interleaves the submodules of $\operatorname{RNEA}$ and $\operatorname{\Delta RNEA}$ together. 
Therefore, the data from each submodule of $\operatorname{RNEA}$ RTP can be directly transferred to the corresponding submodule of $\operatorname{\Delta RNEA}$ RTP. 
The Dynamics Array's submodules can take turns reading data in order, and perform calculations along the array in the form of a pipeline, and then output them in turn. 
This is similar to the behavior of a systolic array.
If there is no need to compute the derivatives, the $\operatorname{\Delta RNEA}$ submodules can be switched to a data pass mode to support standalone operation of the $\operatorname{RNEA}$. 

We can further find that both $\operatorname{RNEA}$ and matrix multiplication are used twice. 
We can further reuse the hardware to perform these two calculations. 
The format of matrix multiplication is unified as shown in Fig.~\ref{Fig.hardware_optimization}c. 
A is a symmetric matrix that can be optimized for sparsity. 

\subsection{Architecture Overview}

\begin{figure}[!tb]
  \centering
  \includegraphics[width=0.98\columnwidth]{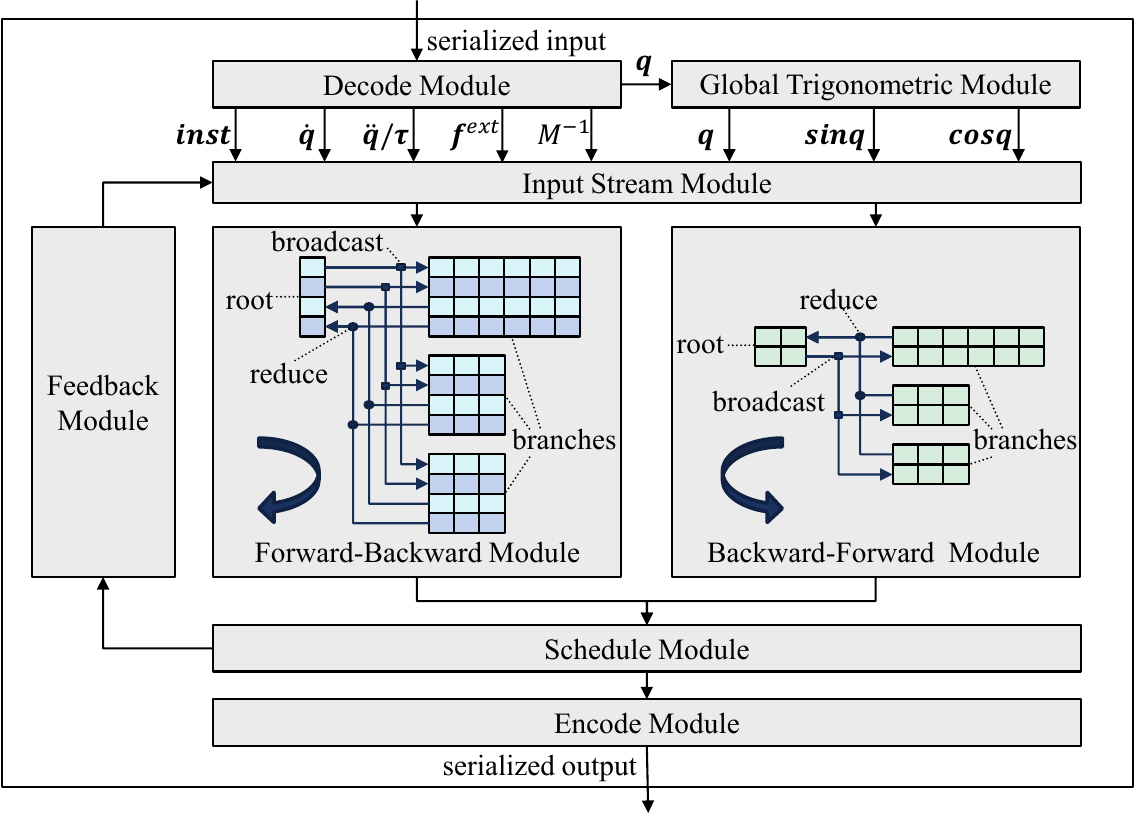}
  \caption{Dadu-RBD architecture.}
  \label{Fig.arch}
\end{figure}

Dadu-RBD needs to be configured according to the model and parameters of the robot before calculation. 
To demonstrate the architecture of Dadu-RBD, we use quadruped robot with an arm (Fig.~\ref{Fig.robot_model}) as an example to configure the accelerator. 
It is a quadruped robot with a mechanical arm, which has $N_B=19$ links, and the DOF of the robot is $N=24$ (including the 6-DOF floating base). 
After the configuration, the structure of Dadu-RBD will be fixed, as shown in Fig.~\ref{Fig.arch}. 

The inputs of Dadu-RBD include $type, q, \dot q, \ddot q/\tau, f^{ext}$ and $M^{-1}$. 
Parameter $type$ indicates which function Dadu-RBD should run. 
It also contains some flag informations to provide function options and output types. 

The outputs of Dadu-RBD contain $\tau, \ddot q, M, M^{-1}, \partial_u \tau$, and $\partial_u \ddot q$.
They can be selected and combined into the output of any function. 
The function $\operatorname{\Delta FD}$ can optionally output $M^{-1}$.

Dadu-RBD contains 8 modules, namely Decode Module, Encode Module, Global Trigonometric Module, Input Stream Module, 
Forward-Backward Module, Backward-Forward Module, Schedule Module and Feedback Module. 
Next, we will briefly introduce the main design methods of the Dadu-RBD architecture. 

\subsubsection{\textbf{Decode and Encode}}
Depending on the chosen function, Dadu-RBD will have different inputs and outputs. 
In order to facilitate the design of the multifunctional pipeline, we unify the formats of all inputs and outputs. 
The Decode Module will deserialized and decode the data from the input interfaces. 
The Encode module can convert data into a CPU-friendly type for subsequent use.

\subsubsection{\textbf{Global Trigonometric Module}}
Most submodules require the values of $sinq$ and $cosq$. 
Therefore we compute all $sinq$ and $cosq$ in advance in Global Trigonometric Module. 
This module computes the Taylor-series expansion of trigonometric functions for fast approximate solutions. 
With loop unrolling optimizations, the entire module can be pipelined to achieve high-throughput performance. 

\subsubsection{\textbf{Scheduling System}}
The scheduling system of Dadu-RBD consists of Input Stream Module, Schedule Module and Feedback Module. 
They constitute a state machine with a feedback structure. 
With the support of this scheduling system, Dadu-RBD can provide timely and accurate data to the multifunctional pipelines, and coordinate conflicts within and between functions. 

The Input Stream Module collects the corresponding data according to the micro-instructions $inst$, and then provides data to the specified module in a certain order. 
There are some subtle differences between the function type $type$ and the micro-instructions $inst$. 
The $type$ is the function interface argument specified by the upper-layer application, while $inst$ is the microinstruction inside Dadu-RBD. 
An instruction $type$ will be translated into multiple different micro-instructions $inst$ during its life cycle, so as to facilitate the scheduling of multifunctional pipelines. 
In this way, modules and submodules can automatically select the output type and path according to the current micro-instructions, thereby realizing the dynamic switching of the dataflow. 

The Schedule Module is responsible for integrating and scheduling the calculation results of multifunctional pipelines. 
It also provides the remaining few vector subtraction and matrix multiplication calculations for the $\operatorname{FD}$, $\operatorname{\Delta FD}$ and $\operatorname{\Delta iFD}$ functions. 
When calculating numerical integration in TO or MPC, Schedule Module can also combine the calculation results of the $\operatorname{FD}$ function with the current state to generate a new integration step.
This new task will be passed to the Feedback Module, waiting for the next calculation.

\subsubsection{\textbf{Forward-Backward and Backward-Forward}}

Forward-Backward and Backward-Forward are the abstraction of the dataflow types in dynamic algorithms.
The dataflow of the Forward-Backward module pass forward first, and then pass backward, 
while the dataflow of the Backward-Forward module is just the opposite. 
Various dynamics algorithms can be implemented through the combination of these two dataflow, only the calculation logic is different. 

The Forward-Backward Module is used to compute $\operatorname{RNEA}$ and its gradient $\operatorname{\Delta RNEA}$. 
It has $4N_B$ submodules, namely $R_{fi}, R_{bi}, D_{fi}$ and $D_{bi}$, as discussed in Section~\ref{rtp-id}. 
The submodules are divided into root and multiple branches according to the structure of the robot. 
The data between them needs to be broadcast and reduced. 
Specifically, we need to copy the data for each branch in the forward pass, and sum the data provided by each branch in the backward pass. 
All submodules in a branch can form a Dynamics Array (Fig.~\ref{Fig.hardware_optimization}b). 

In the current implementation, the Backward-Forward Module is used to compute $\operatorname{MMinvGen}$. 
It has $2N_B+2$ submodules, namely $M_{bi}$ and $M_{fi}$, where $i \in [0,N_B]$, and also be divided into root and multiple branches. 
It has the potential to implement the ABA algorithm, but due to resource constraints we do not currently implement it.

\subsection{Structure-Adaptive Pipelines}

\begin{figure}[!tb]
  \centering
  \subfloat[Mobile arm]
  {\includegraphics[width=.95\columnwidth]{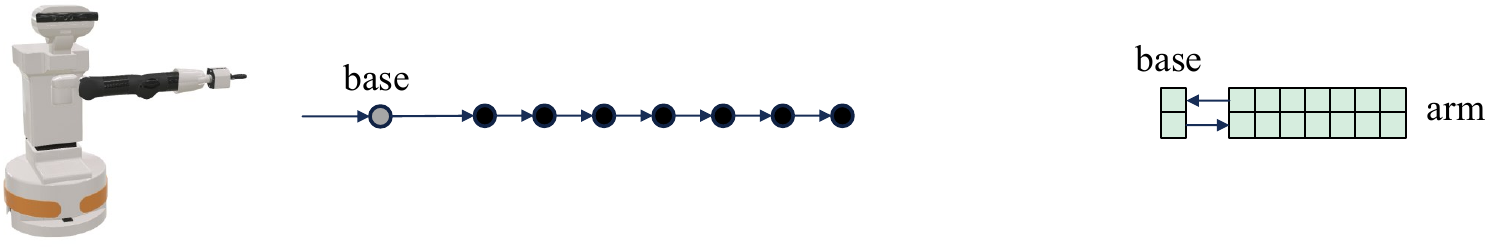}
    \label{Fig.topo_mobile_arm}
  }\\
  \subfloat[Quadruped]
  {\includegraphics[width=.98\columnwidth]{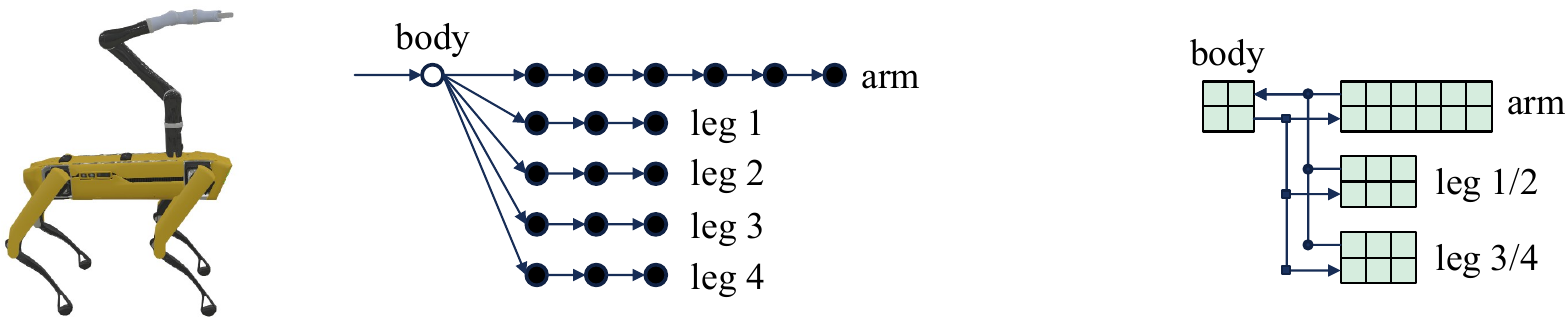}
    \label{Fig.topo_spot_arm}
  }\\
  \subfloat[Humanoid]
  {\includegraphics[width=.98\columnwidth]{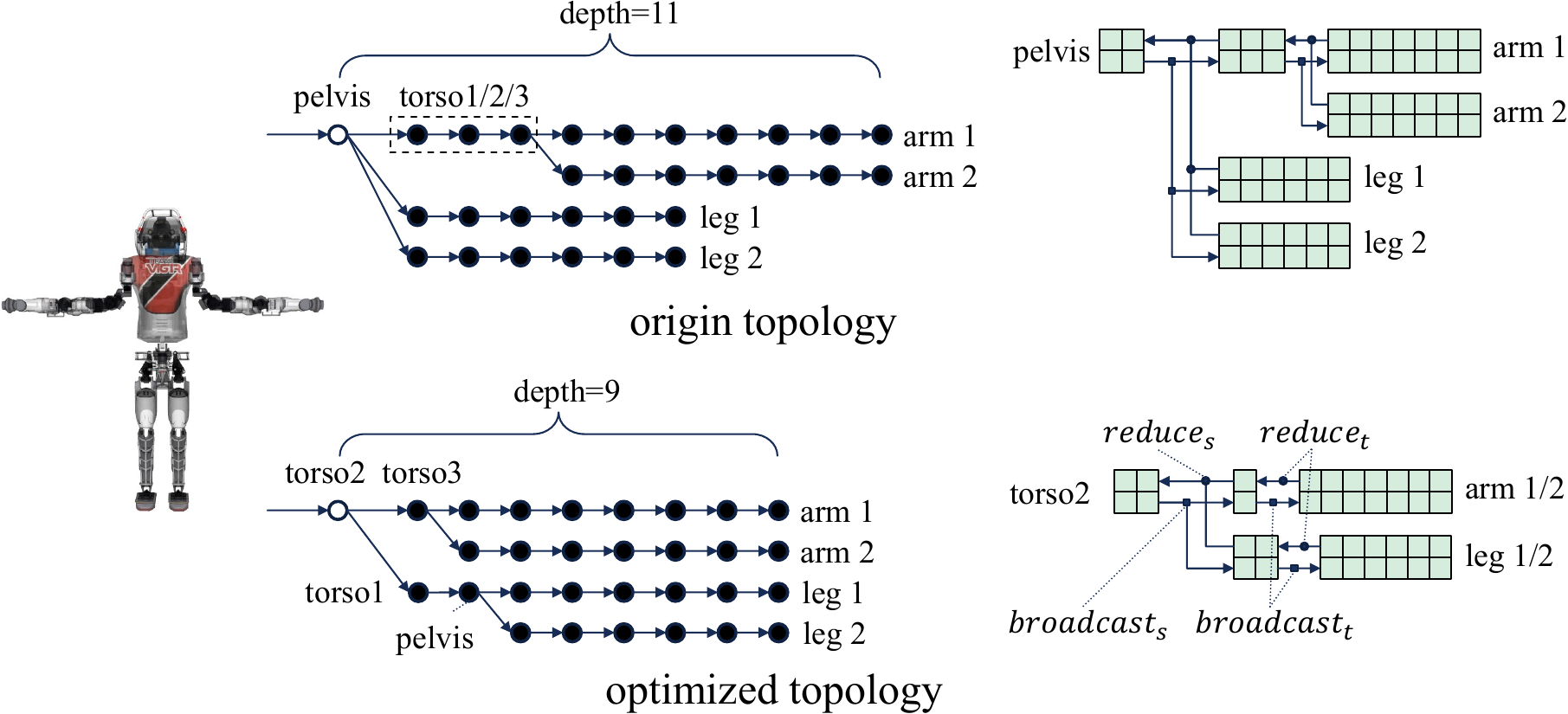}
    \label{Fig.topo_atlas}
  }
  \caption{Optimization for submodules' organization.}
  \label{Fig.struct_arch}
\end{figure}

There are many different structures of robots, 
such as mobile arm (Tiago, Fig.~\ref{Fig.topo_mobile_arm}), 
quadruped robot (Spot-arm, Fig.~\ref{Fig.topo_spot_arm}), 
and humanoid robot (Atlas, Fig.~\ref{Fig.topo_atlas}), etc. 
These different robot structures will change the dependencies between iterations of the dynamics algorithms. 
Therefore, the pipeline structure needs to be adjust according to the robot structure. 
So we call it Structure-Adaptive Pipelines.

\subsubsection{\textbf{Organization of Submodules}}

We design a general optimization method that adapts the organization of Dadu-RBD submodules according to the robot structure. 
We illustrate this approach with the three robot examples above, as shown in Fig.~\ref{Fig.struct_arch}. 
The left side of each subfigure is the topological relationship of the robot structure, and the right side is the organizational relationship of the submodules in Backward-Forward Module. 
The organizational relationship of the submodules in Forward-Backward Module is the same as that of Backward-Forward Module. 

Tiago has a 3-DOF mobile base and a 7-DOF arm, its topology is linear. 
So we can just use a root and a branch to organize all submodules. 
The root joint type is planar, which only has 3 degrees of freedom, so it does not need to be split into two joints. 
As shown in Fig.~\ref{Fig.topo_mobile_arm}, the root corresponds to the base link, and the branch corresponds to the robotic arm. 

Spot-arm has a 6-DOF body, four 3-DOF legs and a 6-DOF arm. 
It has a tree topology. 
We can organize all submodules into a root and three branches. 
As shown in Fig.~\ref{Fig.topo_spot_arm}, the root corresponds to the body link, a 6-size branch corresponds to the robotic arm, and two 3-size branches corresponds to the four legs. 
Here we optimize the number of leg branches. 
Since the legs of the Spot are all symmetrical, the sparsity of the leg parameters is the same, and only a few parameters differ, most of which differ only in sign. 
According to this observation, we can handle two symmetrical legs with one 3-size branch. 
This saves hardware resources. 
At the same time, due to the small topological depth of the leg branches, the complexity of calculating the leg dynamics is much smaller. 
Therefore, the delay of each submodule will be much smaller, and the processing time of one robotic arm can fully support the processing of two legs.

Atlas is a humanoid robot. 
It has a body with a waist, two arms and two legs, forming a topological tree. 
The waist is made up of three joints (torso1, torso2 and torso3) that connect the torso to the pelvis. 
This makes it impossible for branching structures to exist only at the root. 
Traditionally, people will define the pelvis as the root, as shown in Fig.~\ref{Fig.topo_atlas}. 
Under this definition, the depth of the topological tree will become 11. 
As we have discussed before, the complexity of submodules will increase as the iteration depth increases (Section~\ref{Incremental-Calculation}). 
So we can optimize Dadu-RBD by adjusting the depth of robot topology tree, as shown in Fig.~\ref{Fig.topo_atlas}. 
We redefine torso2 as the root of the robot. 
This does not change the robot's topological connectivity, but the depth of the topological tree was reduced to 9, and the depth of each branch is balanced. 
This can reduce a lot of resource consumption, while reducing computing latency. 
Because Atlas is a symmetrical robot, we can also handle two arms or two legs with a single branch array. 

It is worth noting that in Fig.~\ref{Fig.topo_atlas}, the $broadcast_s$ and $reduce_s$ connected to the root (torso2) submodules are oriented to different hardware branches, 
while the $broadcast_t$ and $reduce_t$ connected to torso3 or pelvis submodules are oriented to the same hardware branches, which are time-division multiplexed. 
This difference also exists in the submodules' organization of the Spot-arm robot. 

\subsubsection{\textbf{Pipeline Stages}}

We do not perform fine-grained pipelining for the components of matrix operations, or the resource consumption would be very large. 
We do not only perform coarse-grained pipelining for each module, either. 
Such fine-grained pipelines combined with coarse-grained pipelines are precisely the architecture proposed in the previous work~\cite{Robomorphic}. 
Instead, Dadu-RBD disassembles the algorithm modules into submodules, and then performs medium-grained pipelining among these submodules. 

\begin{figure}[!tb]
  \centering
  \includegraphics[width=.8\columnwidth]{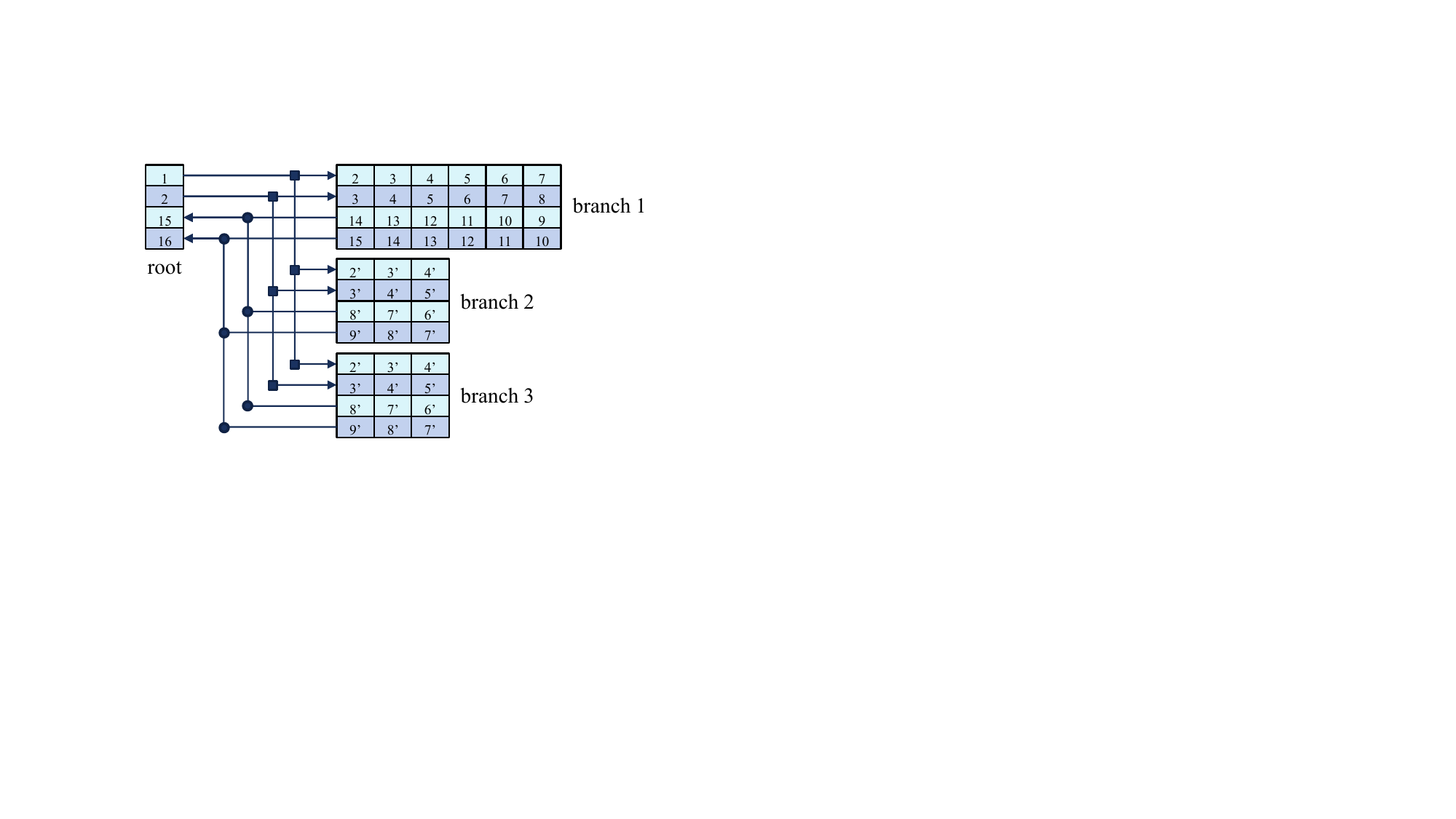}
  \caption{Pipeline Stages for SAPs.}
  \label{Fig.pipeline}
\end{figure}

Fig.~\ref{Fig.pipeline} shows the pipeline stages for the Forward-Backward Module. 
The pipeline of Backward-Forward Module is similar to this. 
Each branch array is independent of each other, so they can perform asynchronous parallel computing.
As the complexity of submodules will increase with the increase of iteration depth (Section~\ref{Incremental-Calculation}), the pipeline cycle of branch 1 is almost double that of branch 2/3. 
So branch 2/3 can handle twice as many tasks as branch 1.

\subsubsection{\textbf{Pipeline Schedule}}

\begin{figure}[!tb]
  \centering
  \includegraphics[width=.95\columnwidth]{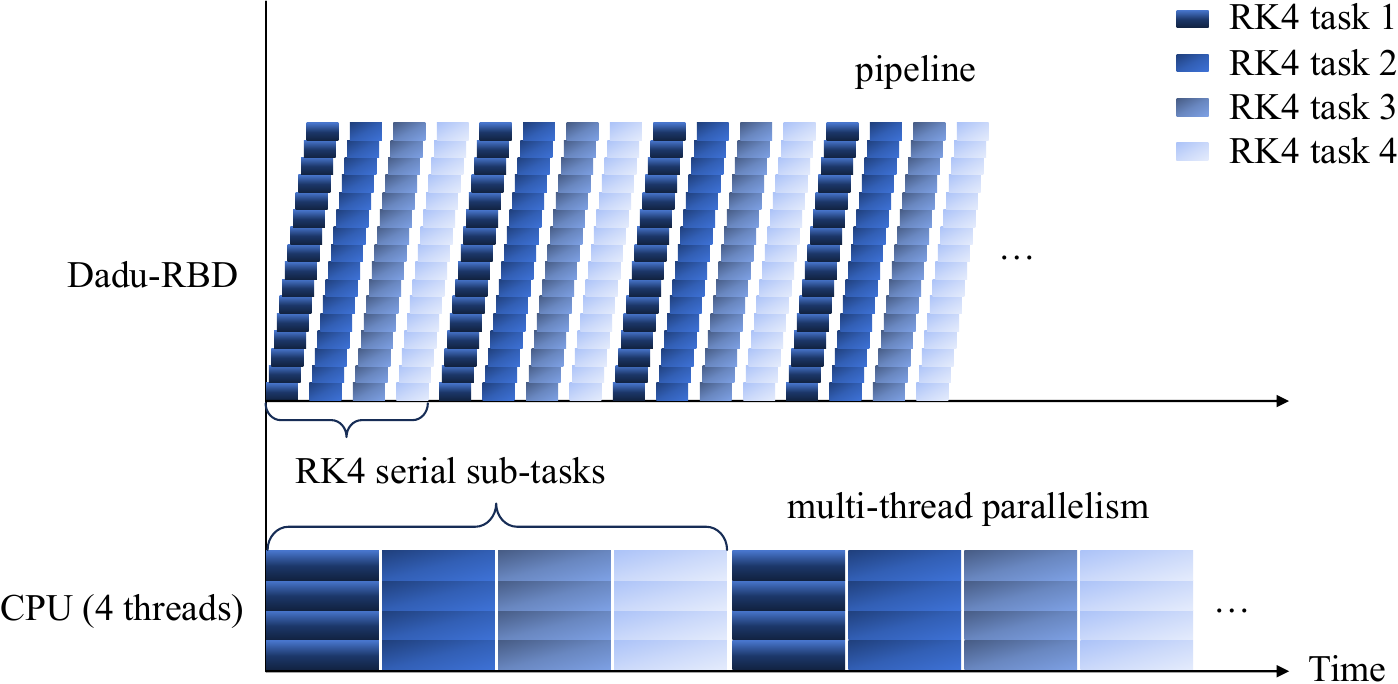}
  \caption{Scheduling method for Dadu-RBD and CPU when processing partially parallelizable tasks. The blocks with the same color are independent tasks.}
  \label{Fig.scheduling_method}
\end{figure}

In many TO and MPC algorithms, computations on multiple sampling points can be processed in parallel. 
We can easily use SPAs to speed up these tasks without worrying about dependencies. 
But sometimes we need to call the dynamics function serially. 
For example, the 4th-order Runge-Kutta sensitivity analysis has 4 serial sub-tasks on each sampling point. 
These tasks need to be scheduled appropriately, as shown in Fig.~\ref{Fig.scheduling_method}. 
Subsequent sub-tasks need to be scheduled after the predecessor tasks are completed. 
Before that, Dadu-RBD can compute other independent batched tasks first. 
Under this scheduling strategy, Dadu-RBD can effectively avoid the negative impact of serial sub-tasks on parallelism. 
For comparison, we also plot how the multithreaded CPU schedules these tasks. 
The CPU performs spatial parallelism through multi-core resources, while Dadu-RBD performs temporal parallelism through SAPs. 

\subsubsection{\textbf{Branch-induced Sparsity}}
In the dynamics algorithms mentioned in this article, the structure of the robot determines the dependencies between variables ($\lambda_i$). 
Therefore, there will be no direct dependencies between submodules of different branches. 
According to this feature, submodules can only keep relevant matrix columns instead of the entire parameter matrix. 
This is robot branch-induced sparsity.

\subsubsection{\textbf{Root Submodules}}
In order to avoid complicated calculations, the base link is split into two parts. 
The robot assumed in the architecture diagram has a floating base with 6 DOF. 
We can split the 6-DOF virtual joint into two 3-DOF virtual joints, which are spherical joint and 3-DOF translation joint. 
This reduces the computational complexity of the root node. 

According to the different needs of the application, Dadu-RBD can provide different modes to handle the dynamics for the root. 
For the virtual joint corresponding to the base link, we can treat it as an ordinary joint and use the standard algorithms for calculation. 
For some bases that are almost not affected by dynamics (for example a robotic arm attached to a very heavy base), we can just provide a state for the base link from the input as an initialization for the following links, or even ignore the root node directly. 
Both Forward-Backward Module and Backward-Forward Module can select the mode of the root submodules through the micro-instructions $inst$. 
This can improve the versatility of the Dadu-RBD accelerator and reduce unnecessary calculations in some cases.

\subsection{Dataflow} \label{dataflow}

\begin{figure}[!tb]
  \centering
  \subfloat[$\operatorname{ID}$]
  {\includegraphics[width=.45\columnwidth]{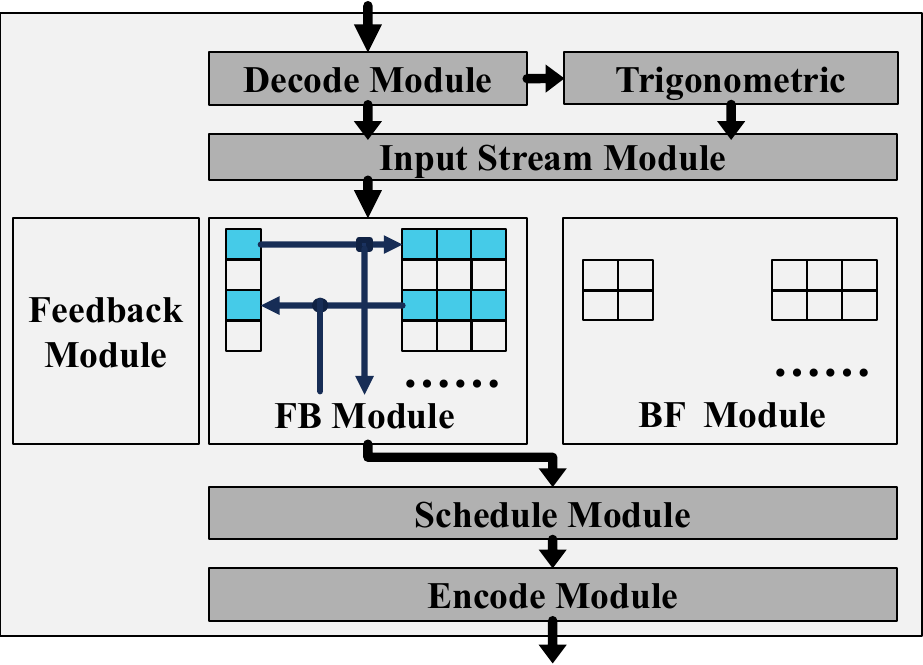}
    \label{Fig.dataflow-ID}
  }\hspace{1pt}
  \subfloat[$\operatorname{FD}$]
  {\includegraphics[width=.45\columnwidth]{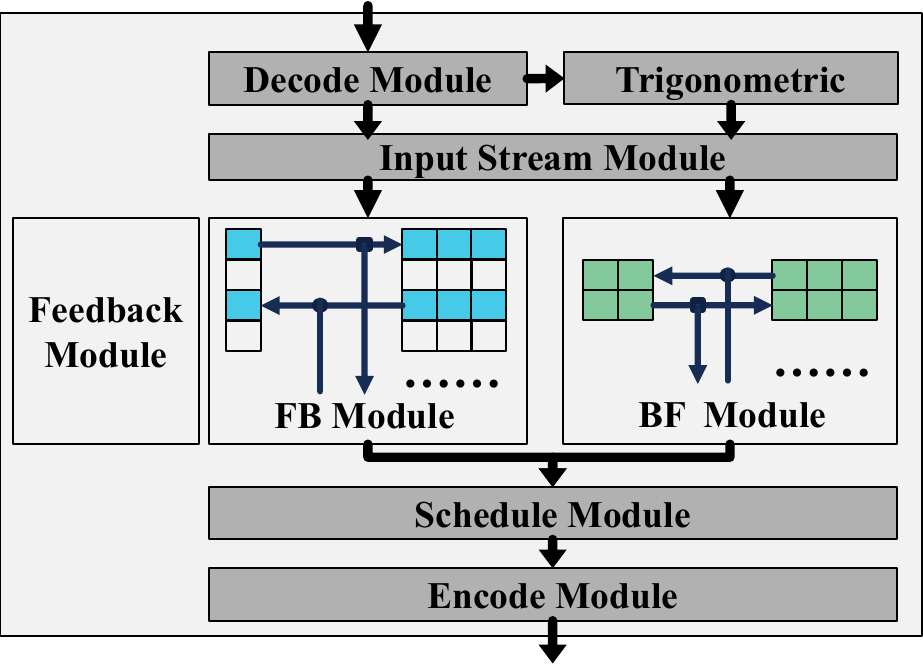}
    \label{Fig.dataflow-FD}
  }\\
  \subfloat[$\operatorname{\Delta ID}$]
  { \includegraphics[width=.45\columnwidth]{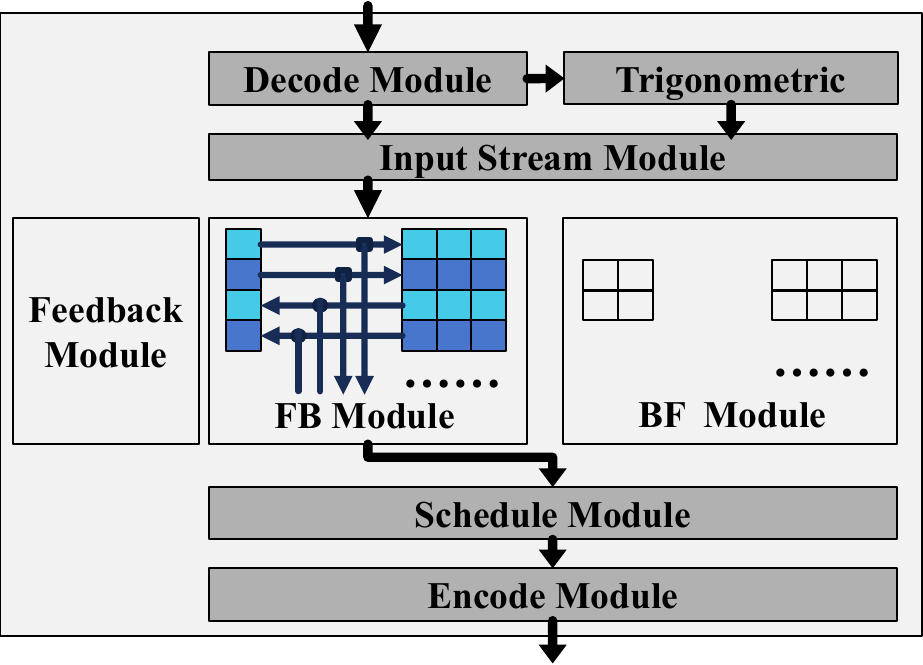}
    \label{Fig.dataflow-dID}
  }\hspace{1pt}
  \subfloat[$\operatorname{M}$ or $\operatorname{Minv}$]
  {\includegraphics[width=.45\columnwidth]{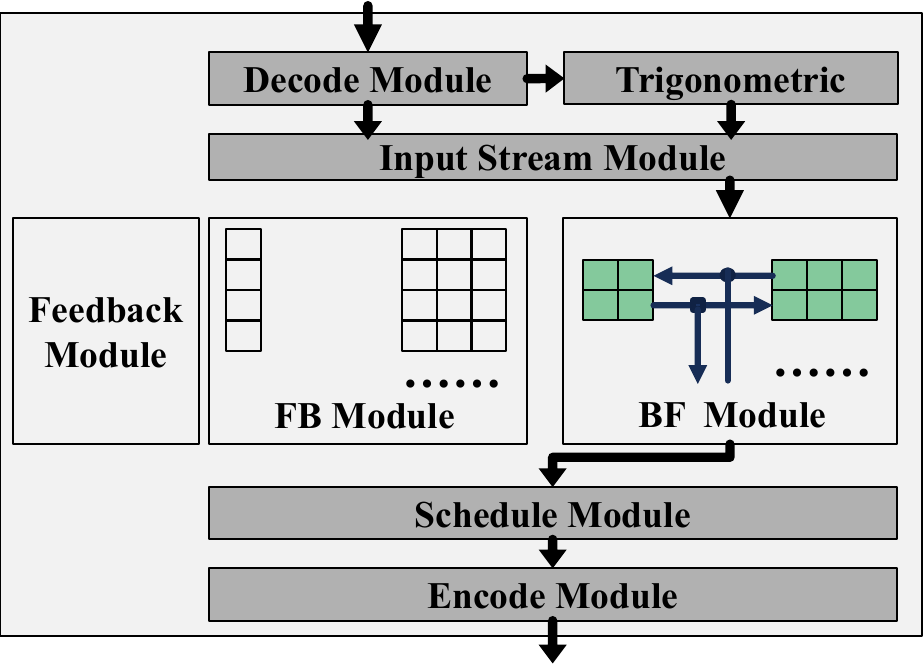}
    \label{Fig.dataflow-Minv}
  }\\
  \subfloat[$\operatorname{\Delta iFD}$]
  {\includegraphics[width=.45\columnwidth]{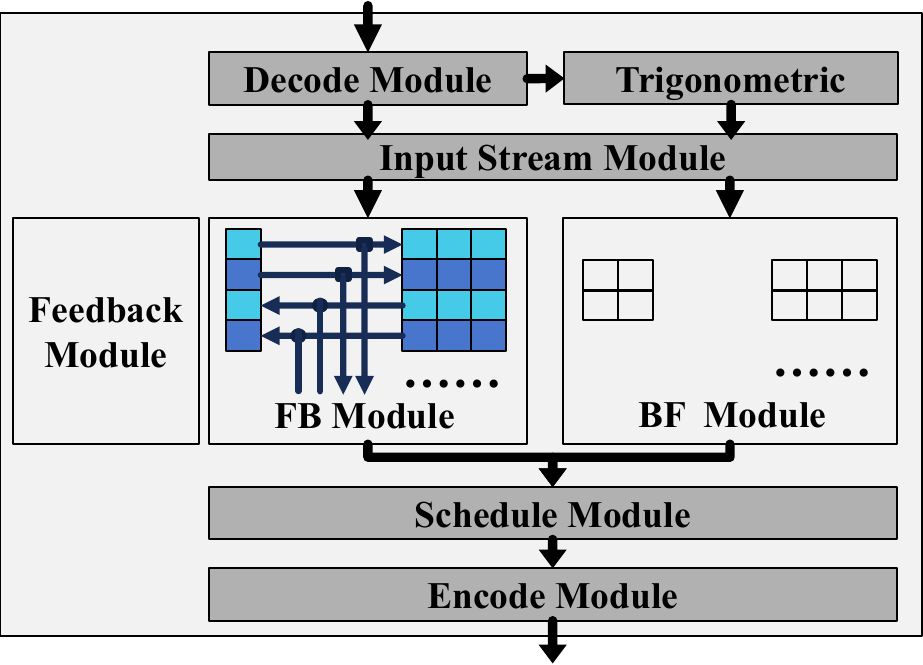}
    \label{Fig.dataflow-diFD}
  }\hspace{1pt}
  \subfloat[$\operatorname{\Delta FD}$]
  {\includegraphics[width=.45\columnwidth]{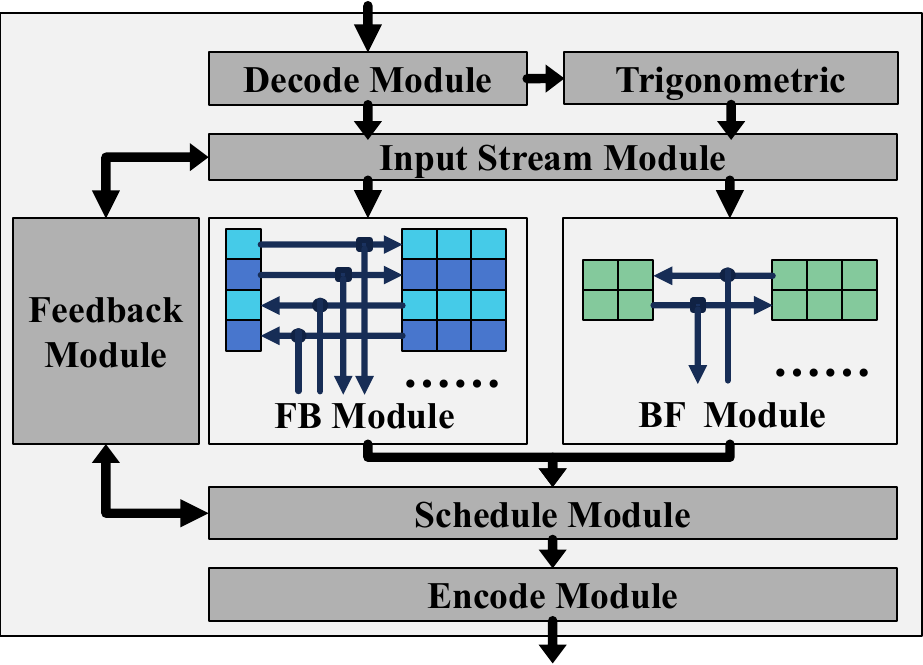}
    \label{Fig.dataflow-dFD}
  }
  \caption{Dataflow of different functions.}
  \label{Fig.dataflow}
\end{figure}

All the modules and submodules described above are completely driven by data, and the data passed between them are implemented by FIFO streams. 
Therefore, these modules and submodules can be organized in the form of dataflow.

As shown in Fig.~\ref{Fig.dataflow}, the dataflow path and activated modules (submodules) for different functions in Table~\ref{table:algo} are marked in dark color. 
For each function, the Decode Module, Global Trigonometric Module, Input Stream Module, Schedule Module and Encode Module all need to be activated.

For functions $\operatorname{ID}$, $\operatorname{FD}$, $\operatorname{M}$, $\operatorname{Minv}$, $\operatorname{\Delta ID}$ and $\operatorname{\Delta iFD}$, 
their dataflow is relatively simple, which is a combination of some modules and submodules of Dadu-RBD. 
While for the function $\operatorname{\Delta FD}$, the dataflow is complex, especially when batch tasks are calculated at the same time. 
The $\operatorname{\Delta FD}$ function has three stages, the first stage is to calculate the $\operatorname{FD}$ function, the second stage is to calculate the $\operatorname{\Delta ID}$ function, and the third stage is to calculate the final result. 
Both the first and second stages use the Forward-Backward Module for calculations, so we need the Feedback Modules to write data back to the Input Stream Module. 
Between stages, the Feedback Module is also responsible for saving necessary intermediate results for a second use, such as the inverse of the mass matrix. 
These different stages are distinguished by the micro-instructions $inst$, so that even with multiple tasks, the overall dataflow will not be messed up.

\section{Evaluation}

We implement and evaluate the Dadu-RBD architecture on the Xilinx platform. 
Vitis and Vivado are used to synthesize the architecture, simulate the dataflow and estimate the power. 
To compare with the widely-used C++ dynamics library Pinocchio \cite{PinocchioRef} and previous state-of-the-art CPU\cite{RBDAcc}, GPU\cite{GRiD_2022} and FPGA\cite{RBDAcc, Robomorphic} works, we use the same FPGA chip as Robomorphic. 
The hardware configurations used in the following evaluations are listed in Table~\ref{table:hardware}. 
Among them, Jetson AGX Orin is NVIDIA's high-performance edge computing module, with 12-core Arm® Cortex®-A78AE CPU and 2048-core NVIDIA Ampere architecture GPU. 
We will use its performance as a baseline for performance evaluation under MAXN Power mode(about 60W). 
In addition, the Core i9-13900HX and RTX 4090 Mobile represent the highest performance levels for mobile CPUs and mobile GPUs, respectively. 
The CPU runtime power can reach up to 140W level, and the GPU can reach up to 175W level. 
They are paired with dual-channel 5600MHz DDR5 memory. 
We will also use their performance as a reference. 
We perform the best compilation optimization options (-O3 and SIMD support) on the CPU and GPU libraries. 

\begin{table}[!tb]
  \centering
  \caption{Hardware Configurations in Evaluations.}
  \label{table:hardware}
  \setlength{\tabcolsep}{1mm}
  \begin{tabular}{|l|l|l|l|l|l|}
    \hline
    \textbf{Type} & \textbf{Processor} & \textbf{Freq} & \textbf{Usage} \\
    \hline
    CPU & AGX Orin & 2.2G & Evaluate \cite{PinocchioRef} \\
    CPU & i9-13900HX & 5.4G & Evaluate \cite{PinocchioRef} \\
    GPU & AGX Orin & 1.3G & Evaluate \cite{GRiD_2022} \\
    GPU & RTX 4090M & 1.8G & Evaluate \cite{GRiD_2022} \\
    FPGA & XVCU9P & 56M & Used in \cite{RBDAcc,Robomorphic} \\
    FPGA & XVCU9P & 125M & Evaluate Dadu-RBD \\
    \hline
  \end{tabular}
\end{table}

We use the Dadu-RBD architecture to realize the dynamics functions for three robots, LBR iiwa \cite{KUKA-LBR-iiwa}, HyQ \cite{doi:10.1177/0959651811402275} and Atlas \cite{AtlasRef}, \nobreak respectively. 
This is consistent with two previous works \cite{PinocchioRef, GRiD_2022}. 
FPGA memory interface bandwidth is limited to a maximum of 32GB/s. 
All evaluations take the I/O time into account. 
The input and output of Dadu-RBD are in the form of data streams, so the I/O overhead of Dadu-RBD can be greatly masked. 
All the results are the averages of millions of runs. 

After experimentation and tuning, Dadu-RBD is able to run at 125MHz in the target FPGA. 
Under this condition, the performance and energy consumption reach a balance.

\subsection{Latency and Throughput}

We compared the latency with CPUs, and compared the throughput with CPUs/GPUs. 
The evaluation method for latency is running 128 different tasks with a single thread, repeated 10k times, and took the average running time of each task. 
Such a task load is to accord with the actual situation. 
The evaluation method for throughput is running 256 batched tasks with multiple threads (12 threads for Jetson AGX Orin, 32 threads for i9-13900HX, grids/blocks/threads for gpu is automatically allocated by library functions), repeated 10k times, and took the average running time of each batched tasks to get the throughput.
256 batched tasks is selected because (1) the highest batch size in previous GPU benchmarks was 256 \cite{GRiD_2022}; 
(2) Referring to the algorithm paper of MPC \cite{MPC-ICHR-2017}, the time horizon of MPC is about 1 second. 
Considering that the frequency requirement of MPC is 100Hz, the corresponding time step is 0.01 second, so the total sampling points are about 100. 
Here we set the number of batches to 256 to cover most of the current needs. 
Of course, we will discuss the impact of larger batch sizes on performance later. 

\begin{figure}[!tb]
  \centering
  \subfloat[iiwa-latency]
  {\includegraphics[width=.45\columnwidth]{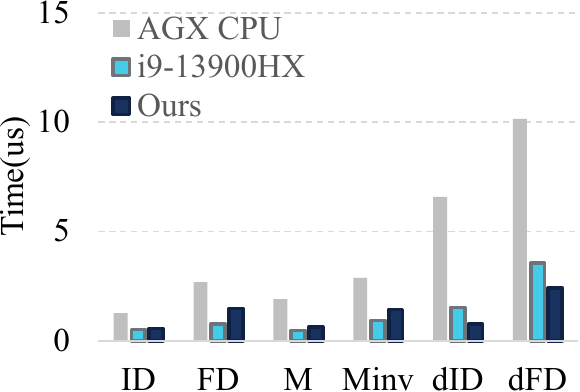}
      \label{Fig.iiwa-lat}
  }\hspace{0pt}
  \subfloat[iiwa-throughput]
  {\includegraphics[width=.45\columnwidth]{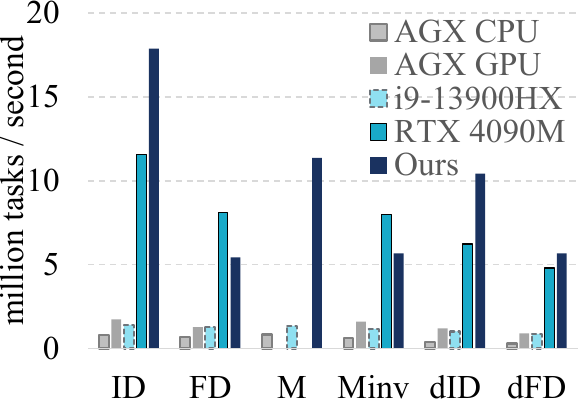}
      \label{Fig.iiwa-thr}
  }\\
  \subfloat[HyQ-latency]
  { \includegraphics[width=.45\columnwidth]{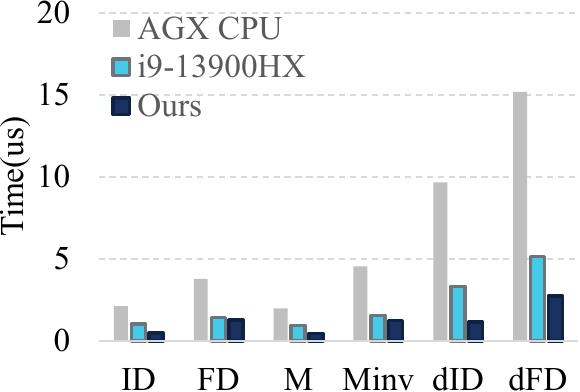}
      \label{Fig.hyq-lat}
  }\hspace{0pt}
  \subfloat[HyQ-throughput]
  {\includegraphics[width=.45\columnwidth]{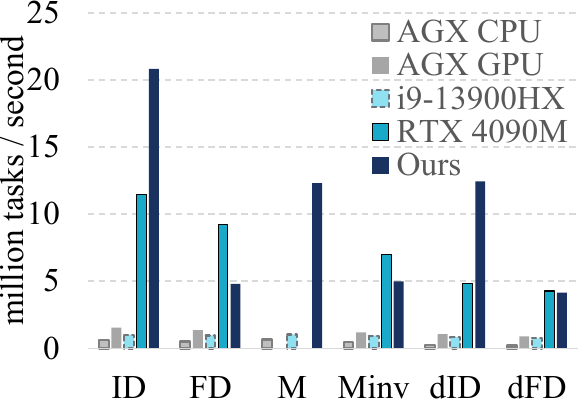}
      \label{Fig.hyq-thr}
  }\\
  \subfloat[Atlas-latency]
  {\includegraphics[width=.45\columnwidth]{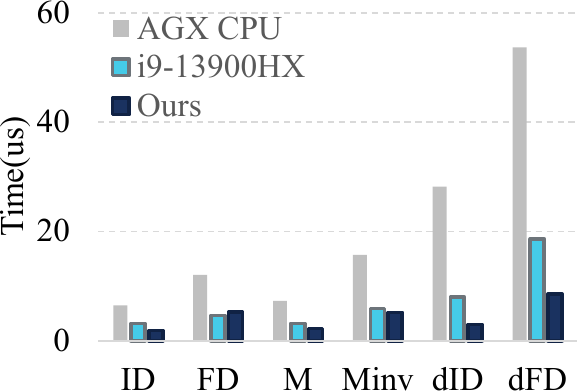}
      \label{Fig.atlas-lat}
  }\hspace{0pt}
  \subfloat[Atlas-throughput]
  { \includegraphics[width=.45\columnwidth]{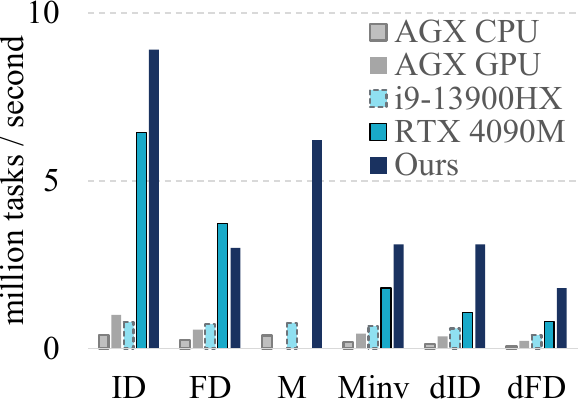}
      \label{Fig.atlas-thr}
  }
  \caption{Latency and throughput comparisons with Pinocchio~\cite{PinocchioRef} on CPU and GRiD \cite{GRiD_2022} on GPU.}
  \label{Fig.performance}
\end{figure}

The performance comparison between Dadu-RBD and CPU, GPU libraries \cite{PinocchioRef,GRiD_2022} are shown in Fig.~\ref{Fig.performance}. 
For all supported functions, the latencies of Dadu-RBD are better than that of the CPU in Jetson AGX Orin. 
When comparing the latency with i9-13900HX, the implementation of Dadu-RBD on the current chip is mostly better than i9, but there are also some functions that are not as good as i9. 

Latencies (lower is better) on different functions against (1) AGX CPU: 0.12$\times$-0.55$\times$, average 0.29$\times$;
(2) i9-13900HX: 0.34$\times$-1.91$\times$, average 0.82$\times$. 

For GPU, GRiD \cite{GRiD_2022} does not realize the calculation of the mass matrix, so there is no data in the corresponding position in the figure. 
The throughput of Dadu-RBD's implementation is much better than that of the AGX CPU, AGX GPU and i9-13900HX. 
But in some functions, it cannot win RTX 4090M. 

Throughput (higher is better) on different functions against (1) AGX CPU: 8.1$\times$-43.6$\times$, average 19.2$\times$; 
(2) AGX GPU: 3.5$\times$-13.4$\times$, average 7.2$\times$;
(3) i9-13900HX: 4.1$\times$-20.2$\times$, average 8.2$\times$;
(4) RTX 4090M: 0.5$\times$-2.8$\times$, average 1.4$\times$.

The acceleration effect of Dadu-RBD on functions $\operatorname{Minv}$ and $\operatorname{FD}$ is not very good, mainly because the inverse of the matrix is really unfriendly to FPGA. 
And in order to save resources, we did not use the most efficient ABA algorithm to calculate $\operatorname{FD}$.

Dadu-RBD is not specifically optimized for latency. Instead, it is designed to maximize the computing throughput. 
So compared with the previous work Robomorphic\cite{Robomorphic}, our computational latency is increased a little bit for the same robot using the same FPGA chip. 
Specifically, the latency of computing $\operatorname{\Delta iFD}$ for robot iiwa is 0.76$\mu$s in Dadu-RBD, while the latency in Robomorphic is 0.61$\mu$s. 
This is mainly due to the additional data transfer overhead caused by the dataflow in Dadu-RBD for supporting multiple functions and the SAPs. 

\begin{figure}[!tb]
  \centering
  \includegraphics[width=.8\columnwidth]{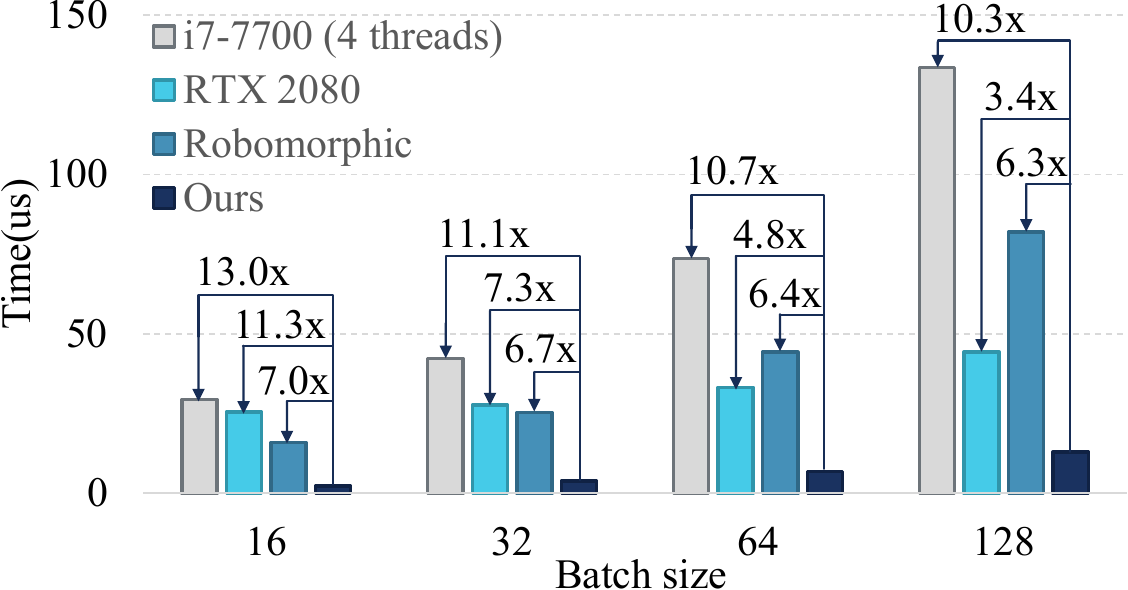}
  \caption{Batched $\operatorname{\Delta iFD}$ tasks compared with CPU, GPU and FPGA.}
  \label{Fig.performance-batch-diFD}\vspace*{-7pt}
\end{figure}

On the other hand, we compare the throughput with the previous works \cite{RBDAcc, Robomorphic}, as shown in Fig.~\ref{Fig.performance-batch-diFD}. (The data is collected from \cite{RBDAcc}.) 
They only implemented the function $\operatorname{\Delta iFD}$ with CPU, GPU and FPGA. 
As a comparison, when running batched tasks, Dadu-RBD further gives 10.3$\times$-13$\times$ performance improvement over their optimized CPU implementation, 
3.4$\times$-11.3$\times$ improvement over their GPU's, 
and 6.3$\times$-7.0$\times$ improvement over their FPGA's.

\begin{figure}[!tb]
  \centering
  \includegraphics[width=.5\columnwidth]{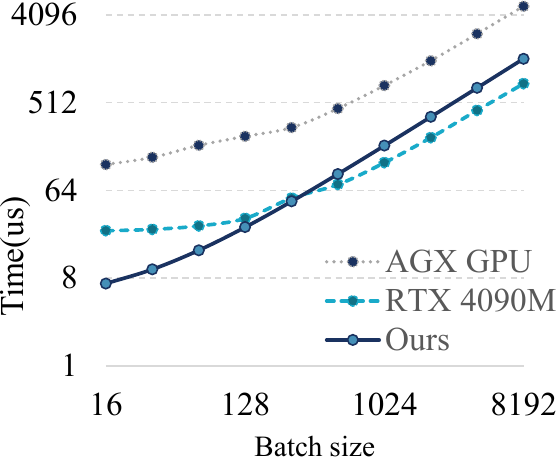}
  \caption{Batched $\operatorname{\Delta FD}$ tasks compared with GPUs.}
  \label{Fig.performance-batch-iFD}\vspace*{-7pt}
\end{figure}

Because the GPU has a strong parallel capability, we also did a larger scale parallel test, as shown in Fig.~\ref{Fig.performance-batch-iFD}. 
It can be seen that the best fit of GPUs is >1024 batch size. The calculation time will increase proportionally and linearly after the size of batches continues to increase. 
RTX 4090M will outperform our implementation when batch size is more than 512. 
Dadu-RBD improves throughput through RTP and SAPs, so the throughput will not fluctuate too much after the pipeline is saturated. 
If we want to further improve throughput, we can instantiate multiple SAPs. 

\subsection{End-to-End Application}

In the robot application mentioned in Fig.~\ref{Fig.robot_example}, there are three kind of tasks can be accelerated by Dadu-RBD, namely \textbf{forward dynamics}, \textbf{inverse of the mass matrix} and \textbf{derivatives of dynamics}, 
which can be calculated by functions $\operatorname{FD}$ and $\operatorname{\Delta FD}$. 
With the help of the RTP and SAPs, Dadu-RBD can achieve 11.2$\times$ performance speedup on these tasks. 
When Dadu-RBD accelerates the supported tasks, the CPU can compute other batch tasks at the same time. 
Compared with a 4-thread CPU implementation, Dadu-RBD can help the entire system increase the control frequency by 80\%.

\subsection{Resource Usage, Power and Energy}

Because each submodule needs to perform computation independently, Dadu-RBD requires many computing resources. 
But compared with the previous work \cite{Robomorphic}, our architecture is able to achieve higher performance with similar resource usage. 
Our implementation takes up 62\% DSP, 17\% FF and 54\% LUT of the target FPGA. 
The previous work mentioned that due to the lack of DSP, they could not instantiate a new independent unit on the same FPGA chip to parallelize the computation, so they also take up at least half of the DSP. 

The power dissipation of our accelerator varies for different functions and robot configurations. 
We estimate Dadu-RBD's power for LBR iiwa. 
For different functions the power dissipation ranges from 6.2W to 36.8W. 
As a comparison with previous work Robomorphic~\cite{Robomorphic} with power 9.6W, which only supports $\operatorname{\Delta iFD}$ function, 
Dadu-RBD's power for function $\operatorname{\Delta iFD}$ is up to 31.2W. 
With 3.25$\times$ higher power, Dadu-RBD's speed is 6.6$\times$ faster than that of Robomorphic, so the energy consumption of Robomorphic is 2.0$\times$ higher than Dadu-RBD's. 
In terms of energy delay product, Dadu-RBD is 13.2$\times$ better then Robomorphic.

\section{Conclusion}

In this paper, by analyzing robot planning and control algorithms, we learned that the calculation of robot dynamics and its derivatives is a basic and important part, and there are often performance bottlenecks. 
Many of the these dynamics functions can be accelerated through our proposed Dadu-RBD architecture. 
By designing RTP/SAPs, the performance and throughput are greatly improved, while keeping resource consumption low. 
The key to the whole work is that we have effectively co-optimized the hardware architecture with the dynamics algorithms and the robot topology structure. 
It is worth noting that there is an independent and concurrent work published on ISCA 2023~\cite{neuman2023roboshape}, which also has similar ideas. 
This idea can also be used in other fields of robotics, waiting for everyone to explore together. 





\bibliographystyle{IEEEtran}
\bibliography{refs}

\begin{thebibliography}{10}
\providecommand{\url}[1]{#1}
\csname url@samestyle\endcsname
\providecommand{\newblock}{\relax}
\providecommand{\bibinfo}[2]{#2}
\providecommand{\BIBentrySTDinterwordspacing}{\spaceskip=0pt\relax}
\providecommand{\BIBentryALTinterwordstretchfactor}{4}
\providecommand{\BIBentryALTinterwordspacing}{\spaceskip=\fontdimen2\font plus
\BIBentryALTinterwordstretchfactor\fontdimen3\font minus
  \fontdimen4\font\relax}
\providecommand{\BIBforeignlanguage}[2]{{%
\expandafter\ifx\csname l@#1\endcsname\relax
\typeout{** WARNING: IEEEtran.bst: No hyphenation pattern has been}%
\typeout{** loaded for the language `#1'. Using the pattern for}%
\typeout{** the default language instead.}%
\else
\language=\csname l@#1\endcsname
\fi
#2}}
\providecommand{\BIBdecl}{\relax}
\BIBdecl

\bibitem{cd-micro16}
S.~Murray, W.~Floyd-Jones, Y.~Qi, G.~Konidaris, and D.~J. Sorin, ``The
  microarchitecture of a real-time robot motion planning accelerator,'' in
  \emph{2016 49th Annual IEEE/ACM International Symposium on Microarchitecture
  (MICRO)}, 2016, pp. 1--12.

\bibitem{DaDu-Series}
Y.~Han, Y.~Yang, X.~Chen, and S.~Lian, ``Dadu series - fast and efficient robot
  accelerators,'' in \emph{2020 IEEE/ACM International Conference On Computer
  Aided Design (ICCAD)}, 2020, pp. 1--8.

\bibitem{Yang_Lian_Chen_Han_2020}
Y.~Yang, S.~Lian, X.~Chen, and Y.~Han, ``Accelerating rrt motion planning using
  tcam,'' in \emph{Proceedings of the 2020 on Great Lakes Symposium on
  VLSI}.\hskip 1em plus 0.5em minus 0.4em\relax Association for Computing
  Machinery, Sep 2020, pp. 481--486.

\bibitem{bakhshalipour2022racod}
M.~Bakhshalipour, S.~B. Ehsani, M.~Qadri, D.~Guri, M.~Likhachev, and P.~B.
  Gibbons, ``Racod: algorithm/hardware co-design for mobile robot path
  planning,'' in \emph{Proceedings of the 49th Annual International Symposium
  on Computer Architecture}, 2022, pp. 597--609.

\bibitem{Jia_Yang_Hsiao_Cruz_Brooks_Wei_Reddi_2022}
\BIBentryALTinterwordspacing
T.~Jia, E.-Y. Yang, Y.-S. Hsiao, J.~Cruz, D.~Brooks, G.-Y. Wei, and V.~J.
  Reddi, \emph{OMU: A Probabilistic 3D Occupancy Mapping Accelerator for
  Real-time OctoMap at the Edge}, May 2022, no. arXiv:2205.03325,
  arXiv:2205.03325 [cs] type: article. [Online]. Available:
  \url{http://arxiv.org/abs/2205.03325}
\BIBentrySTDinterwordspacing

\bibitem{li2020high}
P.~Z.~X. Li, ``High-throughput computation of shannon mutual information on
  chip,'' Ph.D. dissertation, Massachusetts Institute of Technology, 2020.

\bibitem{6386025}
Y.~Tassa, T.~Erez, and E.~Todorov, ``Synthesis and stabilization of complex
  behaviors through online trajectory optimization,'' in \emph{2012 IEEE/RSJ
  International Conference on Intelligent Robots and Systems}, 2012, pp.
  4906--4913.

\bibitem{FastMPC}
M.~Neunert, C.~de~Crousaz, F.~Furrer, M.~Kamel, F.~Farshidian, R.~Siegwart, and
  J.~Buchli, ``Fast nonlinear model predictive control for unified trajectory
  optimization and tracking,'' in \emph{2016 IEEE International Conference on
  Robotics and Automation (ICRA)}, 2016, pp. 1398--1404.

\bibitem{8593840}
M.~Giftthaler, M.~Neunert, M.~Stäuble, J.~Buchli, and M.~Diehl, ``A family of
  iterative gauss-newton shooting methods for nonlinear optimal control,'' in
  \emph{2018 IEEE/RSJ International Conference on Intelligent Robots and
  Systems (IROS)}, 2018, pp. 1--9.

\bibitem{9560742}
E.~Dantec, R.~Budhiraja, A.~Roig, T.~Lembono, G.~Saurel, O.~Stasse,
  P.~Fernbach, S.~Tonneau, S.~Vijayakumar, S.~Calinon, M.~Taix, and N.~Mansard,
  ``Whole body model predictive control with a memory of motion: Experiments on
  a torque-controlled talos,'' in \emph{2021 IEEE International Conference on
  Robotics and Automation (ICRA)}, 2021, pp. 8202--8208.

\bibitem{9783060}
L.~Zhou, M.~Jahnes, and M.~Preindl, ``Modular model predictive control of a
  15-kw, kilo-to-mega-hertz variable-frequency critical-soft-switching
  nonisolated grid-tied inverter with high efficiency,'' \emph{IEEE
  Transactions on Power Electronics}, vol.~37, no.~10, pp. 12\,591--12\,605,
  2022.

\bibitem{Robomorphic}
S.~M. Neuman, B.~Plancher, T.~Bourgeat, T.~Tambe, S.~Devadas, and V.~J. Reddi,
  ``Robomorphic computing: A design methodology for domain-specific
  accelerators parameterized by robot morphology,'' in \emph{Proceedings of the
  26th ACM International Conference on Architectural Support for Programming
  Languages and Operating Systems}, 2021, pp. 674--686.

\bibitem{PinocchioRef}
J.~Carpentier and N.~Mansard, ``{Analytical Derivatives of Rigid Body Dynamics
  Algorithms},'' in \emph{{Robotics: Science and Systems (RSS 2018)}}, Jun.
  2018.

\bibitem{Plancher_Kuindersma_2020}
B.~Plancher and S.~Kuindersma, \emph{\BIBforeignlanguage{en}{A Performance
  Analysis of Parallel Differential Dynamic Programming on a GPU}}, ser.
  Springer Proceedings in Advanced Robotics.\hskip 1em plus 0.5em minus
  0.4em\relax Cham: Springer International Publishing, 2020, vol.~14, pp.
  656--672.

\bibitem{MPC-ICHR-2017}
F.~Farshidian, E.~Jelavic, A.~Satapathy, M.~Giftthaler, and J.~Buchli,
  ``Real-time motion planning of legged robots: A model predictive control
  approach,'' in \emph{2017 IEEE-RAS 17th International Conference on Humanoid
  Robotics (Humanoids)}, 2017, pp. 577--584.

\bibitem{7525066}
D.~Kouzoupis, R.~Quirynen, B.~Houska, and M.~Diehl, ``A block based aladin
  scheme for highly parallelizable direct optimal control,'' in \emph{2016
  American Control Conference (ACC)}, 2016, pp. 1124--1129.

\bibitem{Hereid_Harib_Hartley_Gong_Grizzle_2019}
\BIBentryALTinterwordspacing
A.~Hereid, O.~Harib, R.~Hartley, Y.~Gong, and J.~W. Grizzle, ``Rapid trajectory
  optimization using c-frost with illustration on a cassie-series dynamic
  walking biped,'' no. arXiv:1807.06614, Mar 2019, arXiv:1807.06614 [cs].
  [Online]. Available: \url{http://arxiv.org/abs/1807.06614}
\BIBentrySTDinterwordspacing

\bibitem{arora2009fast}
N.~Arora, R.~P. Russell, and R.~W. Vuduc, ``Fast sensitivity computations for
  trajectory optimization,'' \emph{Advances in the Astronautical Sciences},
  vol. 135, no.~1, pp. 545--560, 2009.

\bibitem{Antony_Grant_2017}
T.~Antony and M.~J. Grant, ``Rapid indirect trajectory optimization on highly
  parallel computing architectures,'' \emph{Journal of Spacecraft and Rockets},
  vol.~54, no.~5, pp. 1081--1091, 2017.

\bibitem{10.1145/3309486.3340246}
Z.~Pan, B.~Ren, and D.~Manocha, ``Gpu-based contact-aware trajectory
  optimization using a smooth force model,'' in \emph{Proceedings of the 18th
  Annual ACM SIGGRAPH/Eurographics Symposium on Computer Animation}, 2019.

\bibitem{Guhathakurta_Rastgar_Sharma_Krishna_Singh_2022}
D.~Guhathakurta, F.~Rastgar, M.~A. Sharma, K.~M. Krishna, and A.~K. Singh,
  ``Fast joint multi-robot trajectory optimization by gpu accelerated batch
  solution of distributed sub-problems,'' \emph{Frontiers in Robotics and AI},
  vol.~9, 2022.

\bibitem{Heinrich_Zoufahl_Rojas_2015}
S.~Heinrich, A.~Zoufahl, and R.~Rojas, ``Real-time trajectory optimization
  under motion uncertainty using a gpu,'' in \emph{2015 IEEE/RSJ International
  Conference on Intelligent Robots and Systems (IROS)}, Sep 2015, pp.
  3572--3577.

\bibitem{Featherstone_2008}
R.~Featherstone, \emph{Rigid Body Dynamics Algorithms}.\hskip 1em plus 0.5em
  minus 0.4em\relax Boston, MA: Springer US, 2008.

\bibitem{Webots}
\BIBentryALTinterwordspacing
Webots, ``http://www.cyberbotics.com,'' open-source Mobile Robot Simulation
  Software. [Online]. Available: \url{http://www.cyberbotics.com}
\BIBentrySTDinterwordspacing

\bibitem{OCS2}
\BIBentryALTinterwordspacing
F.~Farshidian, ``Optimal control for switched systems(ocs2),'' 2018. [Online].
  Available: \url{https://github.com/leggedrobotics/ocs2}
\BIBentrySTDinterwordspacing

\bibitem{6915218}
B.~Käpernick, S.~Süß, E.~Schubert, and K.~Graichen, ``A synthesis strategy
  for nonlinear model predictive controller on fpga,'' in \emph{2014 UKACC
  International Conference on Control (CONTROL)}, 2014, pp. 662--667.

\bibitem{6852127}
F.~Xu, H.~Chen, W.~Jin, and Y.~Xu, ``Fpga implementation of nonlinear model
  predictive control,'' in \emph{The 26th Chinese Control and Decision
  Conference (2014 CCDC)}, 2014, pp. 108--113.

\bibitem{Peyrl_Ferreau_Kouzoupis_2015}
H.~Peyrl, H.~J. Ferreau, and D.~Kouzoupis, ``\BIBforeignlanguage{en}{A hybrid
  hardware implementation for nonlinear model predictive control},''
  \emph{\BIBforeignlanguage{en}{IFAC-PapersOnLine}}, vol.~48, no.~23, pp.
  87--93, Jan 2015.

\bibitem{2464171}
F.~Xu, H.~Chen, X.~Gong, and Q.~Mei, ``Fast nonlinear model predictive control
  on fpga using particle swarm optimization,'' \emph{IEEE Transactions on
  Industrial Electronics}, vol.~63, no.~1, pp. 310--321, 2016.

\bibitem{KHUSAINOV2018105}
B.~Khusainov, E.~Kerrigan, A.~Suardi, and G.~Constantinides, ``Nonlinear
  predictive control on a heterogeneous computing platform,'' \emph{Control
  Engineering Practice}, vol.~78, pp. 105--115, 2018.

\bibitem{MCINERNEY2018381}
I.~McInerney, G.~A. Constantinides, and E.~C. Kerrigan, ``A survey of the
  implementation of linear model predictive control on fpgas,''
  \emph{IFAC-PapersOnLine}, vol.~51, no.~20, pp. 381--387, 2018, 6th IFAC
  Conference on Nonlinear Model Predictive Control NMPC 2018.

\bibitem{RoboX}
J.~Sacks, D.~Mahajan, R.~C. Lawson, and H.~Esmaeilzadeh, ``Robox: An end-to-end
  solution to accelerate autonomous control in robotics,'' in \emph{2018
  ACM/IEEE 45th Annual International Symposium on Computer Architecture
  (ISCA)}, 2018, pp. 479--490.

\bibitem{RBDAcc}
B.~Plancher, S.~M. Neuman, T.~Bourgeat, S.~Kuindersma, S.~Devadas, and V.~J.
  Reddi, ``Accelerating robot dynamics gradients on a cpu, gpu, and fpga,''
  \emph{IEEE Robotics and Automation Letters}, vol.~6, no.~2, pp. 2335--2342,
  2021.

\bibitem{GRiD_2022}
B.~Plancher, S.~M. Neuman, R.~Ghosal, S.~Kuindersma, and V.~J. Reddi, ``Grid:
  Gpu-accelerated rigid body dynamics with analytical gradients,'' in
  \emph{2022 International Conference on Robotics and Automation (ICRA)}, 2022,
  pp. 6253--6260.

\bibitem{45152}
M.~Spong and R.~Ortega, ``On adaptive inverse dynamics control of rigid
  robots,'' \emph{IEEE Transactions on Automatic Control}, vol.~35, no.~1, pp.
  92--95, 1990.

\bibitem{KATAYAMA20206483}
S.~Katayama and T.~Ohtsuka, ``Efficient solution method based on inverse
  dynamics of optimal control problems for fixed-based rigid-body systems,''
  \emph{IFAC-PapersOnLine}, vol.~53, no.~2, pp. 6483--6489, 2020.

\bibitem{doi:10.1177/0278364912469821}
L.~Righetti, J.~Buchli, M.~Mistry, M.~Kalakrishnan, and S.~Schaal, ``Optimal
  distribution of contact forces with inverse-dynamics control,'' \emph{The
  International Journal of Robotics Research}, vol.~32, no.~3, pp. 280--298,
  2013.

\bibitem{tedrake2019drake}
\BIBentryALTinterwordspacing
R.~Tedrake, ``Drake: Model-based design and verification for robotics,'' 2019.
  [Online]. Available: \url{https://drake.mit.edu}
\BIBentrySTDinterwordspacing

\bibitem{control_toolbox}
M.~Giftthaler, M.~Neunert, M.~Stäuble, and J.~Buchli, ``The control toolbox
  — an open-source c++ library for robotics, optimal and model predictive
  control,'' in \emph{2018 IEEE International Conference on Simulation,
  Modeling, and Programming for Autonomous Robots (SIMPAR)}, 2018, pp.
  123--129.

\bibitem{graichen2012real}
K.~Graichen and B.~K{\"a}pernick, \emph{A real-time gradient method for
  nonlinear model predictive control}.\hskip 1em plus 0.5em minus 0.4em\relax
  INTECH Open Access Publisher London, 2012.

\bibitem{Julia_for_robotics}
T.~Koolen and R.~Deits, ``Julia for robotics: simulation and real-time control
  in a high-level programming language,'' in \emph{2019 International
  Conference on Robotics and Automation (ICRA)}, 2019, pp. 604--611.

\bibitem{Crocoddyl}
C.~Mastalli, R.~Budhiraja, W.~Merkt, G.~Saurel, B.~Hammoud, M.~Naveau,
  J.~Carpentier, L.~Righetti, S.~Vijayakumar, and N.~Mansard, ``Crocoddyl: An
  efficient and versatile framework for multi-contact optimal control,'' in
  \emph{2020 IEEE International Conference on Robotics and Automation (ICRA)},
  2020, pp. 2536--2542.

\bibitem{Lynch_Park_2017}
K.~M. Lynch and F.~C. Park, \emph{\BIBforeignlanguage{en}{Modern robotics:
  mechanics, planning, and control}}.\hskip 1em plus 0.5em minus 0.4em\relax
  Cambridge, UK: Cambridge University Press, 2017.

\bibitem{diehl2006fast}
M.~Diehl, H.~G. Bock, H.~Diedam, and P.-B. Wieber, ``Fast direct multiple
  shooting algorithms for optimal robot control,'' \emph{Fast motions in
  biomechanics and robotics: optimization and feedback control}, pp. 65--93,
  2006.

\bibitem{neunert2018whole}
M.~Neunert, M.~St{\"a}uble, M.~Giftthaler, C.~D. Bellicoso, J.~Carius,
  C.~Gehring, M.~Hutter, and J.~Buchli, ``Whole-body nonlinear model predictive
  control through contacts for quadrupeds,'' \emph{IEEE Robotics and Automation
  Letters}, vol.~3, no.~3, pp. 1458--1465, 2018.

\bibitem{6710599}
A.~Krishnamoorthy and D.~Menon, ``Matrix inversion using cholesky
  decomposition,'' in \emph{2013 Signal Processing: Algorithms, Architectures,
  Arrangements, and Applications (SPA)}, 2013, pp. 70--72.

\bibitem{carpentier2018analytical}
\BIBentryALTinterwordspacing
J.~Carpentier, ``{Analytical Inverse of the Joint Space Inertia Matrix},'' May
  2018. [Online]. Available: \url{https://hal.laas.fr/hal-01790934}
\BIBentrySTDinterwordspacing

\bibitem{Istoan_Pasca_2015}
\BIBentryALTinterwordspacing
M.~Istoan and B.~Pasca, ``Fixed-point implementations of the reciprocal, square
  root and reciprocal square root functions,'' Nov 2015. [Online]. Available:
  \url{https://hal.archives-ouvertes.fr/hal-01229538}
\BIBentrySTDinterwordspacing

\bibitem{KUKA-LBR-iiwa}
\BIBentryALTinterwordspacing
K.~AG, ``Lbr iiwa,'' 2022. [Online]. Available:
  \url{https://www.kuka.com/products/robotics-systems/industrial-robots/lbr-iiwa}
\BIBentrySTDinterwordspacing

\bibitem{doi:10.1177/0959651811402275}
C.~Semini, N.~G. Tsagarakis, E.~Guglielmino, M.~Focchi, F.~Cannella, and D.~G.
  Caldwell, ``Design of hyq - a hydraulically and electrically actuated
  quadruped robot,'' \emph{Proceedings of the Institution of Mechanical
  Engineers, Part I: Journal of Systems and Control Engineering}, vol. 225,
  no.~6, pp. 831--849, 2011.

\bibitem{AtlasRef}
\BIBentryALTinterwordspacing
B.~Dynamics, ``Atlas,'' 2022. [Online]. Available:
  \url{https://www.bostondynamics.com/atlas}
\BIBentrySTDinterwordspacing

\bibitem{neuman2023roboshape}
S.~M. Neuman, R.~Ghosal, T.~Bourgeat, B.~Plancher, and V.~J. Reddi,
  ``Roboshape: Using topology patterns to scalably and flexibly deploy
  accelerators across robots,'' in \emph{Proceedings of the 50th Annual
  International Symposium on Computer Architecture}, 2023, pp. 1--13.

\end{thebibliography}










\end{document}